\documentclass{article}

\usepackage[nonatbib]{neurips_2020}

\usepackage[utf8]{inputenc} 
\usepackage[T1]{fontenc}    
\usepackage{hyperref}       
\usepackage{url}            
\usepackage{booktabs}       
\usepackage{amsfonts}       
\usepackage{nicefrac}       
\usepackage{microtype}      
\usepackage{amsmath}
\usepackage{xcolor}
\usepackage{physics}

\usepackage{graphicx}
\usepackage{subfig}

\usepackage{amssymb}
\usepackage{pifont}
\newcommand{\xmark}{\ding{55}}%

\title{Bayesian Neural Ordinary Differential Equations}

\author{%
  Raj Dandekar\\
  Massachusetts Institute of Technology\\
  \texttt{rajd@mit.edu} \\
   \And
   Karen Chung\\
   Massachusetts Institute of Technology\\
   \And
   Vaibhav Dixit \\
   Julia Computing Inc., \\
   Pumas AI\\
   \And
   Mohamed Tarek \\
   University of New South Wales at Canberra, Australia,\\ 
   Pumas AI\\
   \And
   Aslan Garc\'ia-Valadez \\
   National Autonomous University of Mexico.\\
   \And
   Krishna Vishal Vemula\\
   Pramati Technologies\\
   \AND
   Chris Rackauckas \\
   Massachusetts Institute of Technology, \\
   University of Maryland Baltimore,\\
   Pumas AI\\
   \texttt{crackauc@mit.edu} \\
}

\begin{document}

\maketitle

\begin{abstract}

Recently, Neural Ordinary Differential Equations has emerged as a powerful framework for modeling physical simulations without explicitly defining the ODEs governing the system, but instead learning them via machine learning. However, the question: “Can Bayesian learning frameworks be integrated with Neural ODE’s to robustly quantify the uncertainty in the weights of a Neural ODE?” remains unanswered. In an effort to address this question, we primarily evaluate the following categories of inference methods: (a) The No-U-Turn MCMC sampler (NUTS), (b) Stochastic Gradient Hamiltonian Monte Carlo (SGHMC) and (c) Stochastic Langevin Gradient Descent (SGLD). We demonstrate the successful integration of Neural ODEs with the above Bayesian inference frameworks on classical physical systems, as well as on standard machine learning datasets like MNIST, using GPU acceleration. On the MNIST dataset, we achieve a posterior sample accuracy of 98.5\% on the test ensemble  of 10,000 images. Subsequently, for the first time, we demonstrate the successful integration of variational inference with normalizing flows and Neural ODEs, leading to a powerful Bayesian Neural ODE object. Finally, considering a predator-prey model and an epidemiological system, we demonstrate the probabilistic identification of model specification in partially-described dynamical systems using universal ordinary differential equations. Together, this gives a scientific machine learning tool for probabilistic estimation of epistemic uncertainties.
\end{abstract}

\section{Introduction}
The underlying scientific laws describing the physical world around us are often prescribed in terms of ordinary differential equations (ODEs). Recently, Neural Ordinary Differential Equations \cite{chen2018neural} has emerged as a powerful framework for modeling physical simulations without explicitly defining the ODEs governing the system, but learning them via machine learning. By noticing that in the limit of infinite layers, a ResNet module \cite{he2016deep} behaves as a continuous time ODE, Neural ODEs allow the coupling of neural networks as expressive function transformations, and powerful purpose built ODE solvers. While \cite{chen2018neural} explored a number of applications of the Neural ODE framework, their success in a Bayesian inference framework remains unexplored. 

Simultaneously, there has been an emergence of efficient Bayesian inference methods suited for high-dimensional parameter systems, such as the No-U-Turn MCMC sampler (NUTS) \cite{hoffman2014no} which is an extension of the Hamiltonian Monte Carlo Algorithm, and Stochastic Gradient Markov Chain Monte Carlo (SGMCMC) methods like Stochastic Gradient Hamiltonian Monte Carlo (SGHMC) \cite{chen2014stochastic} and Stochastic Gradient Langevin Descent (SGLD) \cite{welling2011bayesian}.

A number of works in literature explored the use of Bayesian methods to infer parameters of systems defined by ODEs \cite{Lunn2002, RevModPhys.83.943, GIROLAMI20084, Huang2020} and others used Bayesian methods to infer parameters of neural network models, e.g. \cite{Jospin2020HandsonBN, DBLP:journals/corr/abs-1902-02476, DBLP:journals/corr/abs-1907-07504}. Bayesian neural networks in particular has been an active area of research for a while. The readers are referred to the excellent recent tutorial by \cite{Jospin2020HandsonBN} for an overview of recent advances in the field. However, this prompts the question: “Can Bayesian learning frameworks be integrated with Neural ODE’s to robustly quantify the uncertainty in the weights of a Neural ODE?” 

In an effort to address this question, we demonstrate and compare the integration of Neural ODEs with the following methods of Bayesian Inference: (a) The No-U-Turn MCMC sampler (NUTS), (b) Stochastic Gradient Hamiltonian Monte Carlo (SGHMC) and (c) Stochastic Langevin Gradient Descent (SGLD). We present successful results on classical physical systems and on standard machine learning datasets (using GPU acceleration); especially on the standard MNIST dataset, we achieve a test ensemble accuracy of $98.5 \%$ on $10000$ images. This is a performance competitive with current state-of-the-art image classification methods, which meanwhile lack our method's ability to quantify the confidence in its predictions. Subsequently, we premiere the integration of Bayesian Neural ODEs with variational inference, the predictive power of which improves with the introduction of normalizing flow.

Finally, advancing from learning a physical system's differential equations via Bayesian Neural ODEs, we consider the problem of recovering missing terms from a dynamical system using universal differential equations (UDEs) \cite{rackauckas2020universal}. Using the Preconditioned SGLD variation of SGLD, we demonstrate the predictive success of Bayesian UDEs on (a) a predator-prey model and (b) the epidemiological model of COVID-19 spread. Through this, we present a viable method for the probabilistic quantification of epistemic uncertainties via a hybrid machine-learning and mechanistic-model-based technique.

Our approach differs from that of \cite{andreas2019differential} who mainly looked at integration of Bayesian methods with Neural SDE's, and not Neural ODEs describing physical systems or large scale deep learning datasets like the MNIST dataset, which we consider here.

In this study, we used the Julia differentiable programming stack \cite{rackauckas2020generalized} to compose the Julia differential equation solvers \cite{rackauckas2017differentialequations} with the Turing probabilistic programming language \cite{ge2018t, xu2019advancedhmc}. The study was performed without modifications to the underlying libraries due to the composability afforded by the differentiable programming stack.
\section{Results}

We illustrate the robustness of the Bayesian Neural ODE framework through the following case studies: \newline

\textbf{Case study 1: Spiral ODE} \newline

The Spiral ODE model is prescribed by the following system of equations:
\begin{align}\label{spiral}
    \frac{du_{1}}{dt} & = - \alpha u_{1}^{3} + \beta u_{2}^{3}\\
    \frac{du_{2}}{dt} & = - \beta u_{1}^{3} - \alpha u_{2}^{3}
\end{align}

\textbf{Case study 2: Lotka-Volterra ODE} \newline

The Lotka-Volterra predator-prey model is prescribed by the following system of equations:
\begin{align}\label{lv}
    \frac{du_{1}}{dt} & = - \alpha u_{1} - \beta u_{1} u_{2}\\
    \frac{du_{2}}{dt} & = - \delta u_{2} + \gamma u_{1} u_{2}
\end{align}

\subsection{Bayesian Neural ODE: NUTS Sampler}
The No-U-Turn-Sampler (NUTS) is an extension of the Hamiltonian Monte Carlo (HMC) algorithm. Through a recursive algorithm, NUTS automatically determines when the sampler should stop an iteration, and thus prevents the need to specify user defined parameters, like the number of steps $L$. In addition, through a dual averaging algorithm, NUTS adapts the step size $\epsilon$ throughout the sampling process. \newline
We define the parameters of the $d$ dimensional Neural ODE by $\theta$. The action of the Neural ODE on an input value $u_{0}$ generates an output $\Tilde{Y} = \textrm{NNODE}_{\theta} (u_{0})$. The input data is denoted by $\hat{Y}$. The loss function, $L$ is defined as

\begin{equation}\label{loss}
    L(\theta) = \sum_{i=1}^{d} || \hat{Y_{i}} - \tilde{Y_{i}}||^{2} 
\end{equation}

The model variables $\theta$ and the momentum variables $r$ are drawn from the joint distribution 
\begin{equation}\label{potential}
    p (\theta, r) \propto \textrm{exp} [\mathcal{L(\theta)} - \frac{1}{2} r.r]
\end{equation}

where $\mathcal{L}$ is the logarithm of the joint density of $\theta$. In terms of a physical analogy, if $\theta$ denotes a particle's position, then $\mathcal{L}$ can also be viewed as the negative of the potential energy function and $\frac{1}{2} r.r$ denotes the kinetic energy of the particle.\newline

In the Bayesian Neural ODE framework, we define $\mathcal{L}(\theta)$ as 
\begin{equation}
    \mathcal{L}(\theta) = -\sum_{i=1}^{d} || \hat{Y_{i}} - \tilde{Y_{i}}||^{2} -  \theta . \theta
\end{equation}

The $\theta . \theta$ term indicates the use of Gaussian priors. We adapt the step size of the leapfrog integrator using Nesterov's dual averaging algorithm \cite{hoffman2014no} with $\delta$ as the target acceptance rate. Finally, we define the number of warmup samples as $n_{w}$ and the number of posterior samples collected as $n_{p}$.

\begin{figure}
\centering
\begin{tabular}{cc}
\subfloat[]{\includegraphics[width=0.45\textwidth]{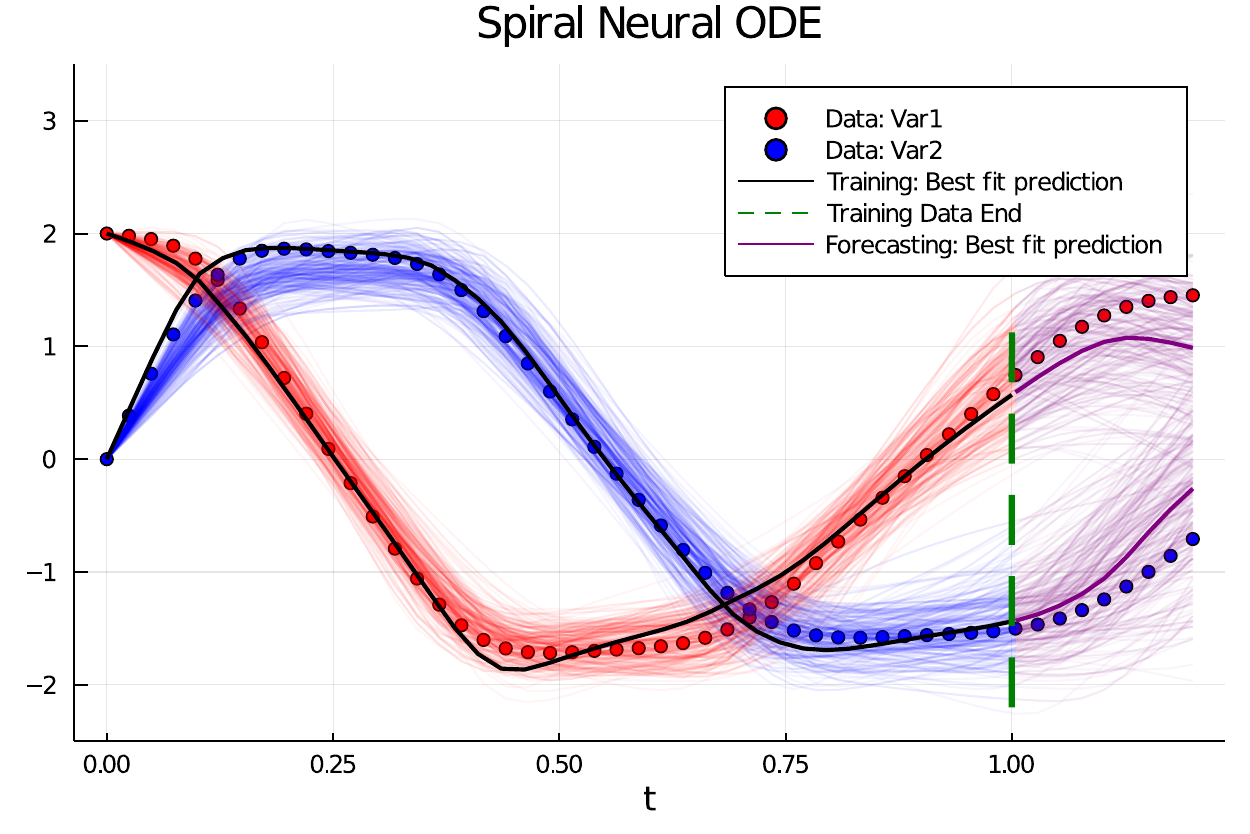}}
\subfloat[]{\includegraphics[width=0.45\textwidth]{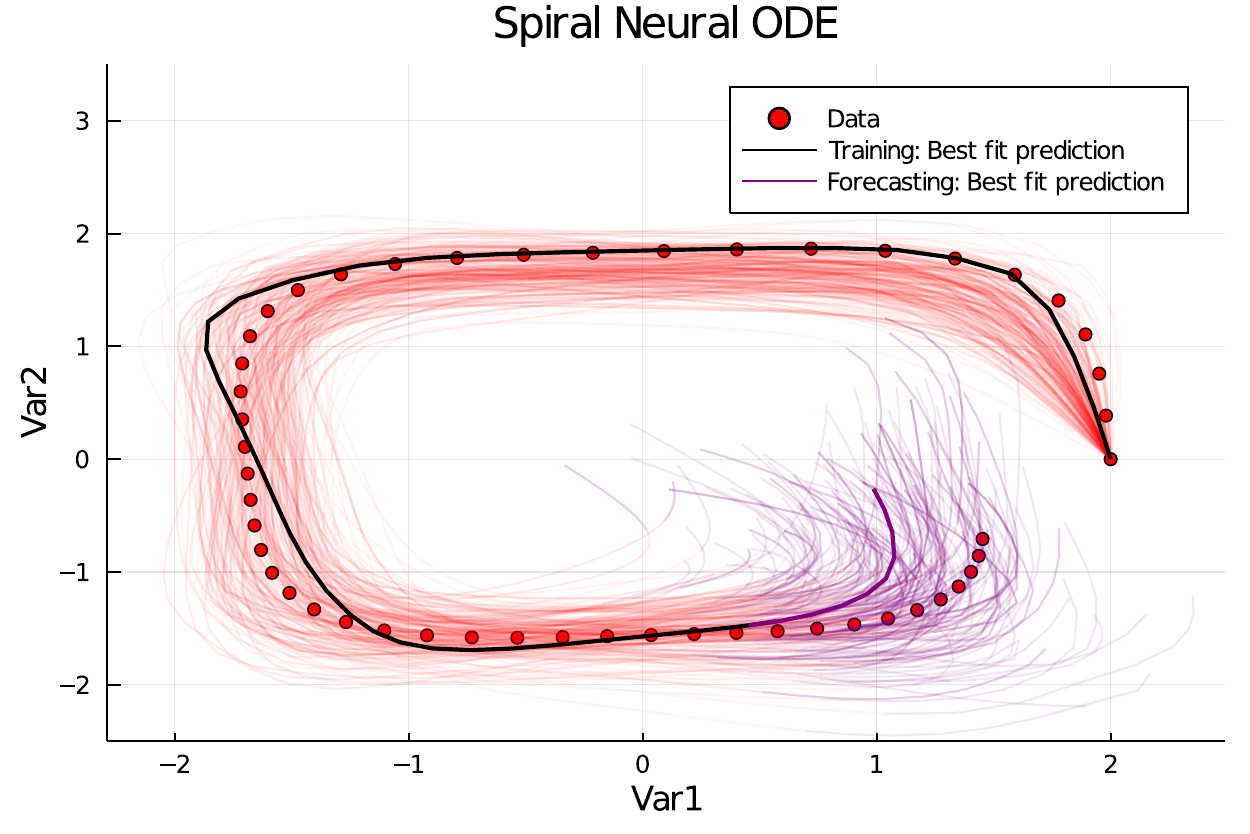}}\\
\subfloat[]{\includegraphics[width=0.45\textwidth]{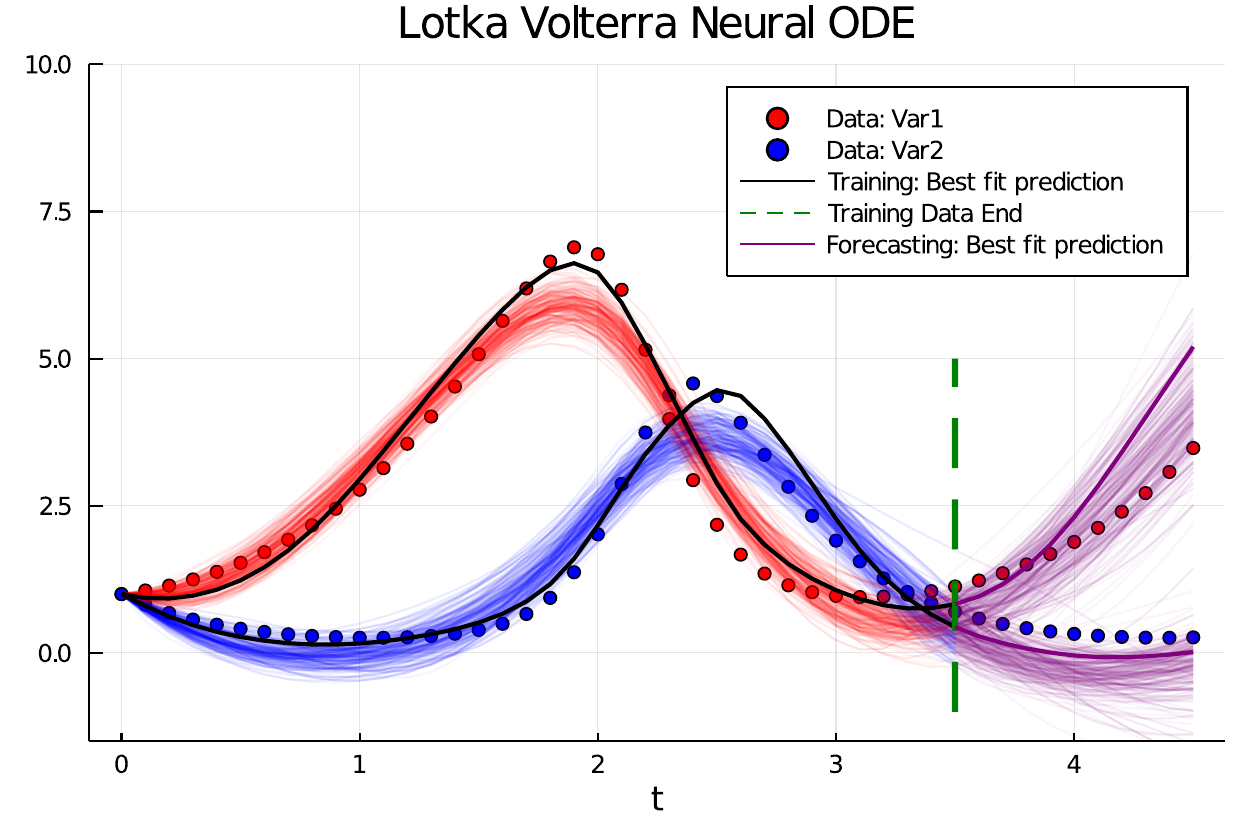}}
\subfloat[]{\includegraphics[width=0.45\textwidth]{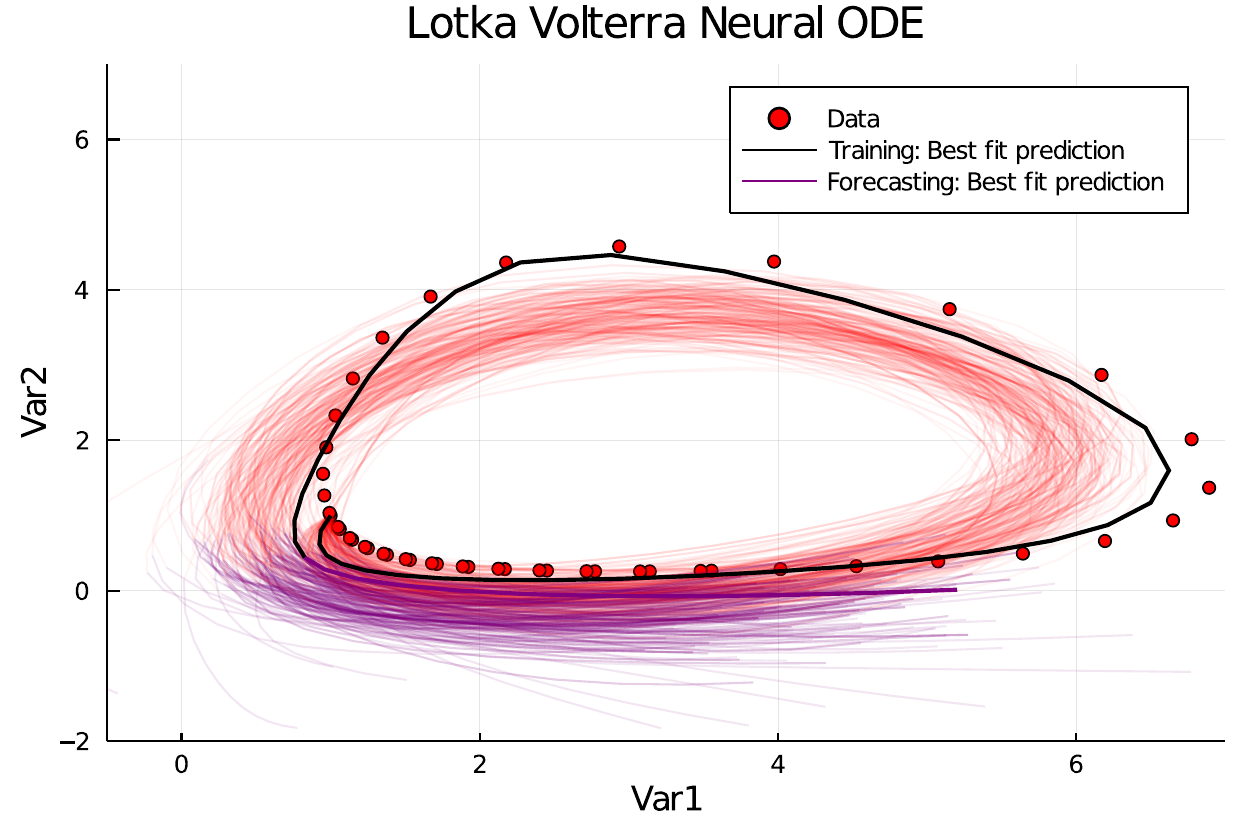}}
\end{tabular}
\caption{Comparison of the Bayesian Neural ODE: NUTS prediction and estimation compared with data for (a,b) Case study 1 and (c,d) Case study 2.}\label{spiralfig}
\end{figure}

\begin{figure}
\centering
\begin{tabular}{cc}
\subfloat[]{\includegraphics[width=0.45\textwidth]{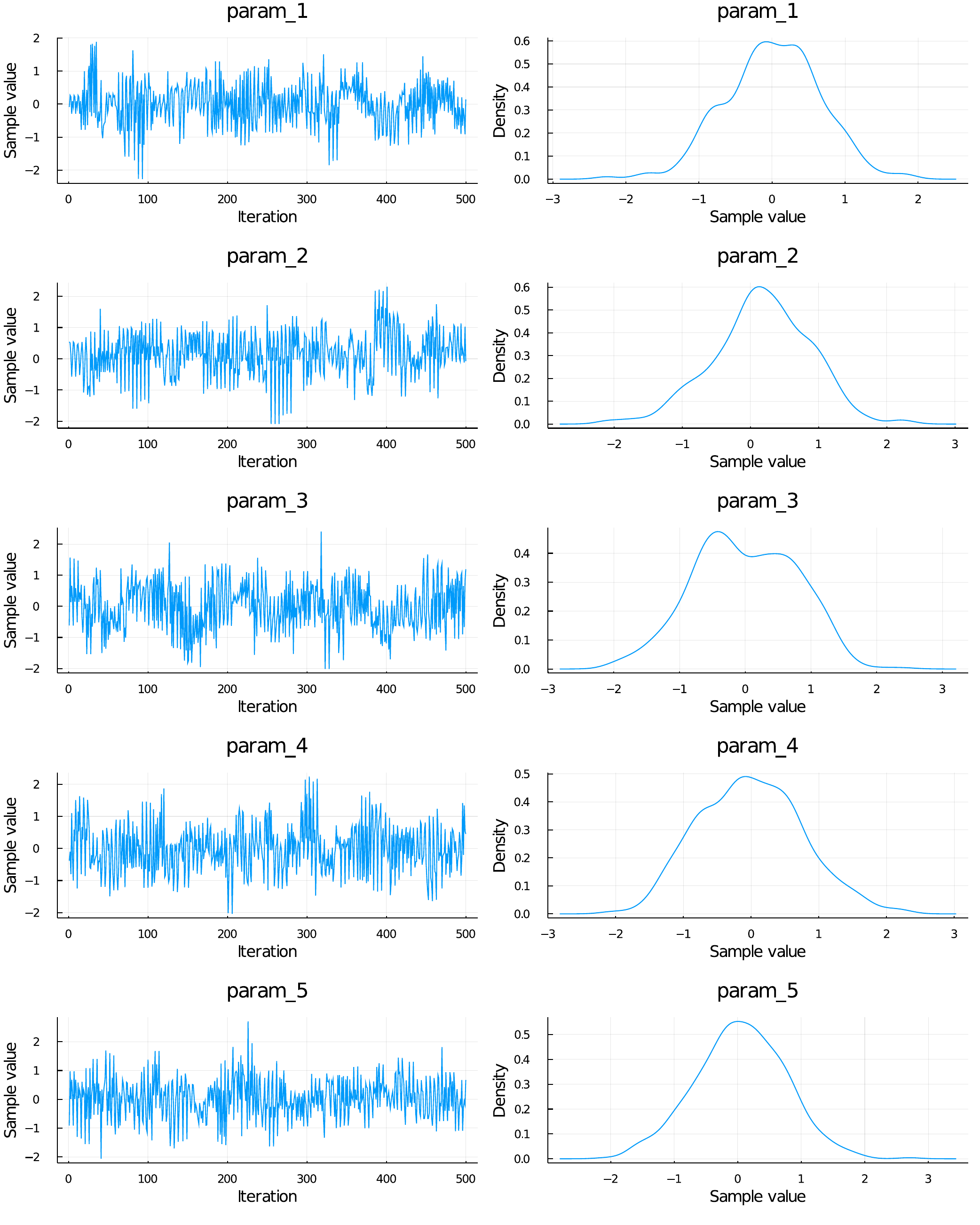}}
\subfloat[]{\includegraphics[width=0.23\textwidth]{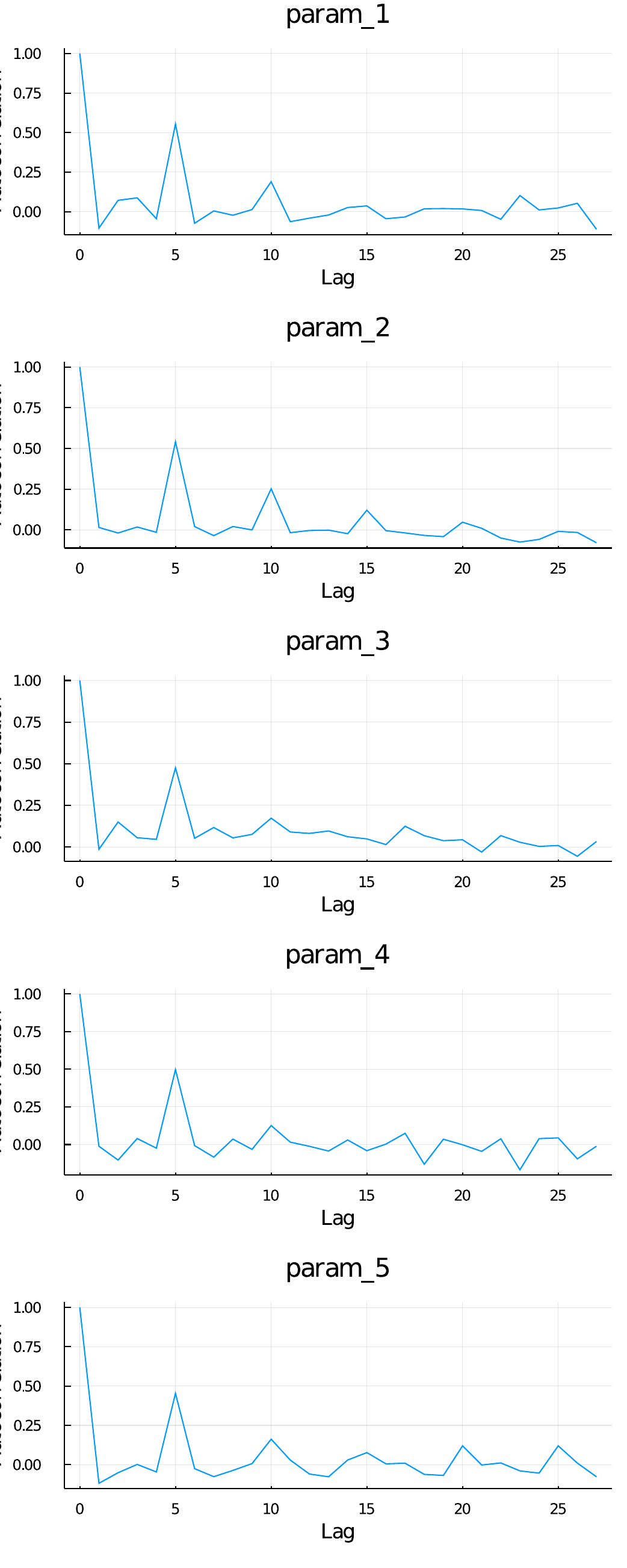}}\\
\end{tabular}
\caption{For the Spiral ODE example (Equations 1-2), using the NUTS framework, figure shows: (a) Trace plots and Density plots of the posterior and (b) auto-corelation plot for the first 5 parameters.}\label{spiral_diagnostics}
\end{figure}

We apply the Bayesian Neural ODE framework outlined above to case studies 1 and 2 given in Equations 1-4. For case study 1, we use $\alpha = 0.1, \beta = 2$. In the NUTS algorithm, we use $\delta = 0.45, n_{w} = 1000, n_{s} = 500$. 
For case study 2, we use $\alpha = 1.5, \beta = 1, \gamma = 3, \delta = 1$; with $\delta = 0.45, n_{w} = 500, n_{s} = 1000$. $2$ layers with $50$ units in each layer and $\textrm{tanh}$ activation function was used as the neural ODE architecture for both examples. \newline

From figure \ref{spiralfig}, we can see that the Bayesian Neural ODE: NUTS prediction and forecasting for both case studies outlined in Equations 1-4 are consistent with the ground truth data. \newline

Figure \ref{spiral_diagnostics}a showing the posterior density and trace plots for the first 5 parameters of the Spiral ODE example, shows that the samples are well mixed. The quick decay seen in the auto-corelation plot shown in figure \ref{spiral_diagnostics}b also indicates a fast mixing Markov chain. This is also confirmed by the effective sample size extracted for the posterior chain of $500$ samples, which shows values of $362, 470, 134, 509, 661$ for the first 5 parameters. Similar well mixed plots are seen for all parameters, but are not shown here for the sake of brevity.

\begin{table}
  \caption{Spiral ODE: Effect of NUTS acceptance ratio and Neural ODE architecture. Number of warmup samples, $n_{w} = 500$ and number of posterior samples, $n_{s} = 500$ for all cases shown. The minimum loss value obtain is similar in all cases shown.}
  \label{table_nuts}
  \centering
  \begin{tabular}{llll}
    \toprule
    \cmidrule(r){1-2}
    $\delta$     & Units & Layers & Time (s)  \\
    \midrule
    0.45 & 5 & 2 & 480 \\
    0.45  & 10 & 2  & 900  \\
    0.45  & 50 & 2  & 2100  \\
    0.45  & 100 & 2  & 3900  \\
    0.45   & 10 & 3 & 3300  \\
    0.45  & 10 & 4  & 10200  \\
    0.65   & 50 & 2  & 3400 \\
    0.85  & 50 & 2 & 7900  \\
    0.95  & 50 & 2 & 8600 \\
    \bottomrule
  \end{tabular}
\end{table}

Table \ref{table_nuts} shows the effect of NUTS acceptance ratio and Neural ODE architecture on the Bayesian Neural ODE performance for the Spiral ODE example. The minimum loss value obtain is similar in all cases shown. Thus, we see that even the smallest neural architecture with $2$ layers and $5$ units in each layer gives the optimal loss performance, and with a considerably better timing performance. Among different NUTS acceptance ratios ($\delta$) tested, the best timing performance is given by the lowest acceptance ratio, $\delta = 0.45$.

\subsection{Bayesian Neural ODE: SGHMC}
The Stochastic Gradient Hamiltonian Monte Carlo Sampler (SGHMC) is a method that combines HMC's effective state space exploration with stochastic gradient methods' computational efficiencies \cite{chen2014stochastic}.
SGHMC injects friction to the "momentum" auxiliary variables that parameterize the target distribution's Hamiltonian dynamics. 

The Hamiltonian function $H(\theta, r) = U(\theta) + \frac{1}{2}r^TM^{-1}r$ measures the total "energy" of a system with position variables $\theta$ and momentum variables $r$. The potential energy function is given by $U = - \sum_{x \in \mathcal{D}}^{}\log p(x | \theta) - \log p(\theta)$; mass matrix $M$ and $r$ define the kinetic energy term.  

To sample from the posterior distribution $p(\theta | \mathcal{D})$, HMC considers generating samples from the joint distribution $\pi(\theta, r)$ $\propto$ exp($H(\theta, r))$, proposing samples according to the Hamiltonian dynamics:
\begin{align}\label{hmc}
    d\theta = M^{-1}r dt \\
    dr = -\nabla U(\theta) dt
\end{align} 
Here, SGHMC adds a "friction" term to the momentum update. In practice, we consider $\nabla\tilde{U}$, a noisy estimate of $\nabla U$. Similarly, $\hat{B}$ is defined as an estimate of $B(\theta) = \frac{1}{2}\epsilon V(\theta)$, the diffusion matrix contributed by the covariance of the stochastic gradient noise $V(\theta)$. SGHMC additionally introduces a user-specified friction term $C \succeq \hat{B}$. In total, by defining $v = \epsilon M^{-1}r$, $\eta = \epsilon^2 M^{-1}$, $\alpha = \epsilon M^{-1}C$, and $\hat{\beta} = \epsilon M^{-1}\hat{B}$ for stepsize $\epsilon$, SGHMC iteratively updates the variables for sampling according to: 
\begin{align}\label{sghmc}
    \theta_{t} &:= \theta_{t} + \Delta{\theta_{t}} \\
    v_{t} &:= v_{t} + \Delta{v_{t}} \\
    \Delta\theta_{t} &:= v \\
    \Delta v_t &:= -\eta\nabla\tilde{U}(x) - \alpha v + \mathcal{N}(0, 2(\alpha - \hat{\beta})\eta)
\end{align}
 Naturally, $\eta$ corresponds to the learning rate and $\alpha$ the momentum decay. 
 
We apply the Bayesian Neural ODE framework outlined above to case studies 1 and 2 given in Equations 1-4. For case study 1, we use  $\alpha$ = 1, $\beta$ = 1. In the SGHMC algorithm, we use $\eta$ = $1.5^{-6}$ and $\alpha$ = 0.07 and draw 2500 posterior samples. We define a prior distribution centered at the MAP point with a standard deviation of 0.4. For case study 2, we use $\alpha$ = 1.5, $\beta$ = 1.0, $\gamma$ = 3.0, and $\delta$ = 1.0; we draw 350 samples with SGHMC using hyperparameters $\eta = 7.0^{-6}$ and $\alpha = 0.07$. We define a prior distribution centered at the MAP point with a standard deviation of 1.0. For both case studies, the neural ODE architecture consists of 2 layers with tanh activation function; there are 50 units in each layer for case study 1, and 10 units for case study 2.   

\begin{figure}
\centering
\begin{tabular}{cc}
\subfloat[]{\includegraphics[width=0.36\textwidth]{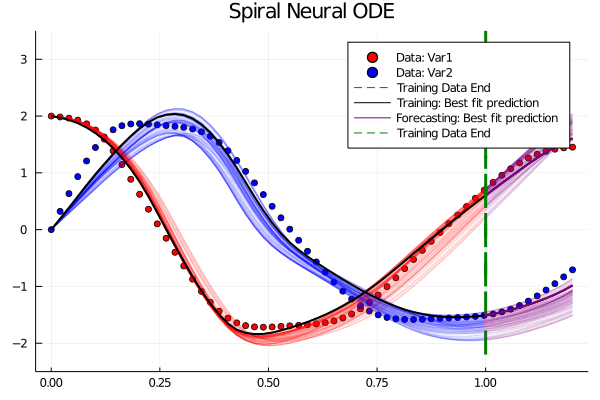}}
\subfloat[]{\includegraphics[width=0.36\textwidth]{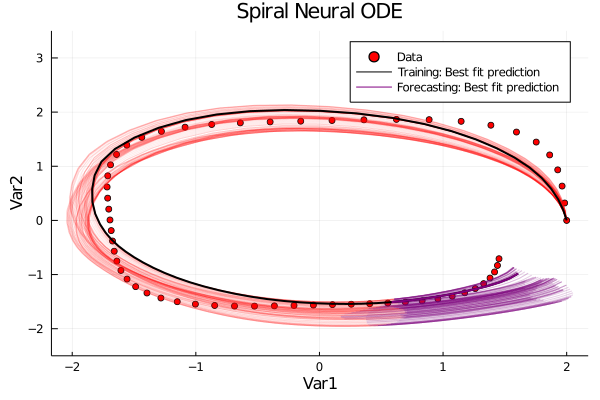}}\\
\subfloat[]{\includegraphics[width=0.36\textwidth]{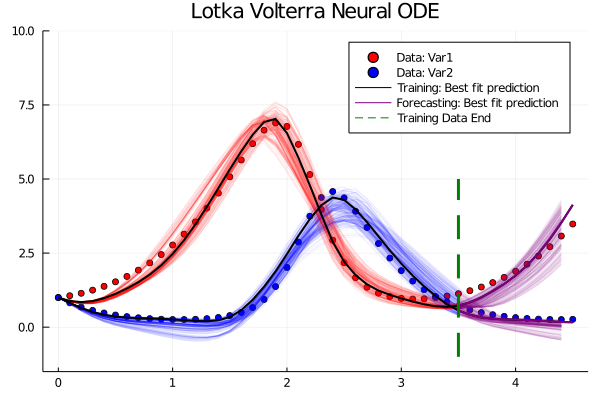}}
\subfloat[]{\includegraphics[width=0.36\textwidth]{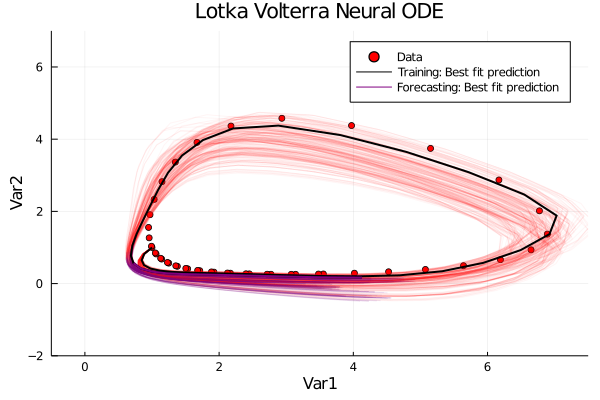}}
\end{tabular}
\caption{Comparison of Bayesian Neural ODE SGHMC's prediction and forecasting against ground truth data for (a,b) case study 1 and (c,d) case study 2.}\label{spiralfig_sghmc}
\end{figure}

Figure \ref{spiralfig_sghmc} illustrates the consistency of Bayesian Neural ODE SGHMC's prediction and forecasting with ground truth data, for both case studies (Equations 1-4).

\subsubsection{SGHMC on the MNIST dataset}

\begin{figure}[h]
\centering
\begin{tabular}{c}
\subfloat[]{\includegraphics[width=0.74\textwidth]{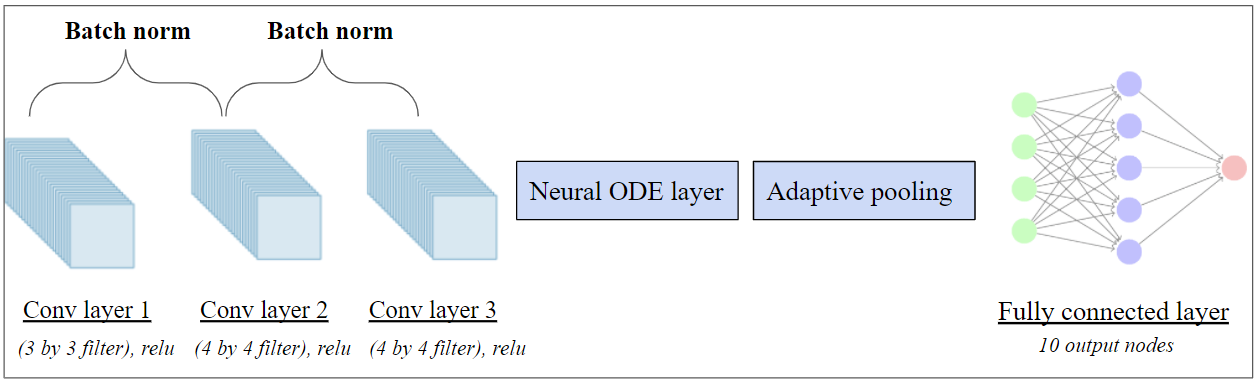}}
\end{tabular}
\caption{Neural network architecture used for the image classification task on MNIST. The Neural ODE contains two convolutional layers. The network has 208010 parameters in total. This architecture was combined with the SGHMC method to lead to a Bayesian Neural ODE object which can be used for image classification.}\label{architecturefig}
\end{figure}

\begin{table}[h]
\centering
\caption{Performance on MNIST using the Bayesian Neural ODE: SGHMC approach is outlined in the present study. Here, 310 posterior samples for each image in a test set of 10,000 images is considered. The best fit test error represents the mean of the number of erroneous predictions in all the posterior samples ($310$) for all images ($10000$). }
\begin{tabular}{cccccc}
\toprule
& Error & Neural ODE? & Best fit   &Reference  \\ & estimates? & & test error & test error & \\ \midrule
RK-Net                & \xmark & \checkmark    &  0.47 $\%$               & Chen (2018)           \\
ODE-Net             & \xmark & \checkmark      &  0.42 $\%$                 & Chen (2018)            \\
Bayesian Alex-Net             & \checkmark & \xmark      &   1 $\%$             & Shridhar (2019) \\
Bayesian LeNet-5             & \checkmark & \xmark       &   2 $\%$                & Shridhar (2019)          \\

Bayesian Neural ODE (ensemble)      & \checkmark & \checkmark  &   0.78 $\%$                & Our study      \\

\end{tabular}\label{MNIST1}
\end{table}

We now apply Bayesian Neural ODE with SGHMC to a image classification task on the MNIST dataset. A ResNet \cite{resnet2015} layer behaves as a continuous time ODE at the limit of infinite layers. Given this natural analogy, we here implement ODE layers in place of the residual layers used in classic image recognition architectures. 

Specifically, we design three convolutional layers interspersed with Batch Normalization layers: the initial layer has a $3 \times 3$ filter and the next two convolutional layers each have a $4 \times 4$ filter (stride $2 \times 2$, padding $1 \times 1$, ReLU activations). A Neural ODE layer with two convolutional layers with $3 \times 3$ filter (padding 0,  ReLU activations) is appended to act as a residual layer. Finally, we add an adapative average pooling layer and a fully connected layer consisting of $10$ neurons (one per class). The full architecture is visualized in figure \ref{architecturefig}.

Given the Neural ODE architecture, we initialize SGHMC with decay schedule of $\epsilon_{t} = \eta . t^{-\gamma}$ and parameters $\gamma = 0.01$, $\eta = 0.5$ and $\alpha = 0.1$, execute for $2530$ iterations, and sample the last $310$ parameter updates. We find that tempering—scaling the standard deviation of the added noise in SGHMC by a constant \cite{li2015preconditioned}—significantly reduces time for convergence. Our experiment uses $10^4$ as the tempering constant. SGHMC is more computationally expensive, requiring ten epochs through the training dataset, than a simple MAP estimation; a trial MAP optimization with ADAM yields a $98.7 \%$ test accuracy after a single-epoch run. 

\begin{figure}[h]
\centering
\begin{tabular}{c}
\subfloat[]{\includegraphics[width=0.5\textwidth]{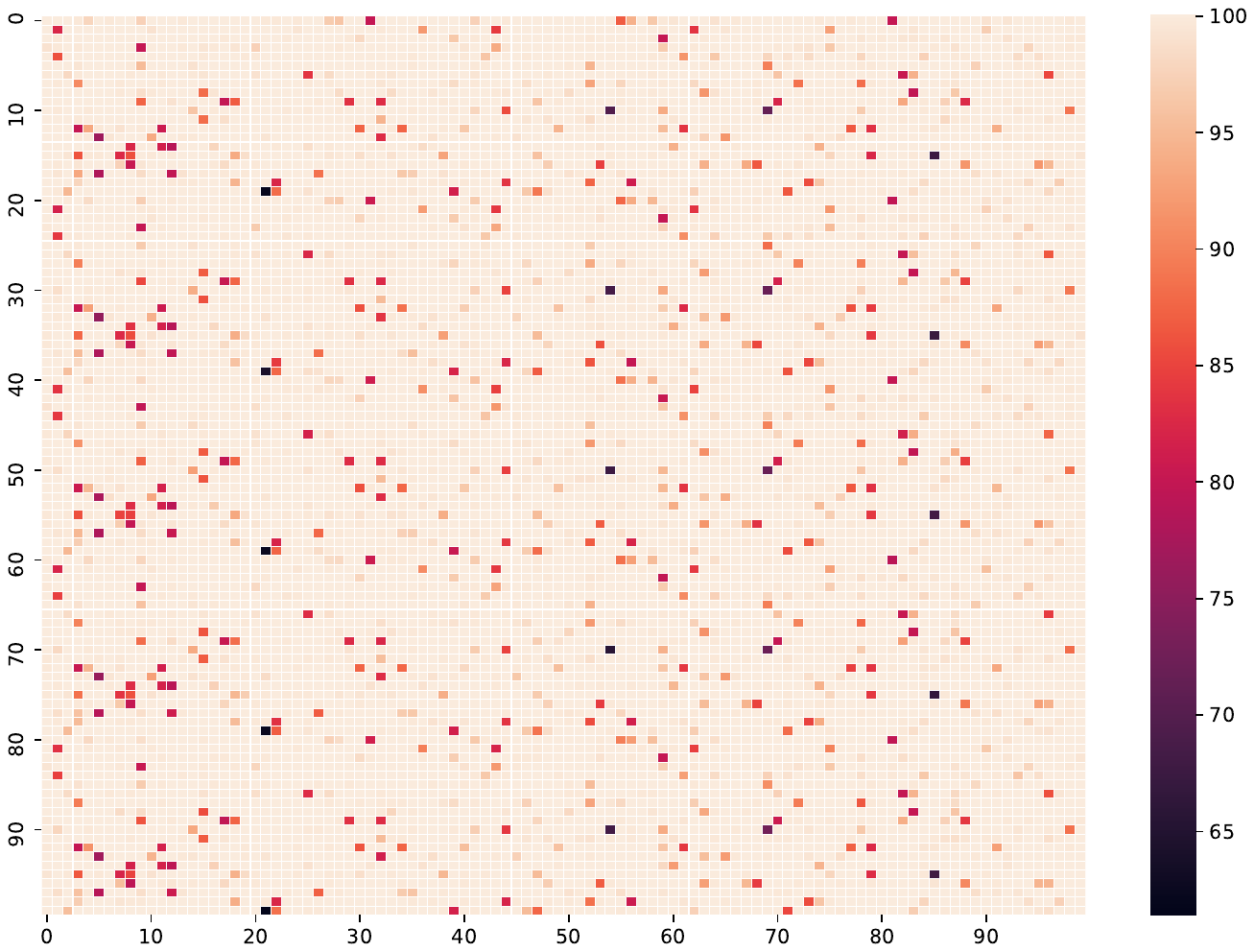}}
\end{tabular}
\caption{Bayesian Neural ODE with SGHMC is applied to the MNIST dataset. Each cell in this figure represents the percentage of correct predictions out of $310$ posterior samples on a single image. Results for the entire test set of $10,000$ images is visualized here as a $100 \times 100$ heatmap}\label{MNISTfig1}
\end{figure}

The test set consists of all 10,000 images in the MNIST dataset. Each cell in the heatmap of figure \ref{MNISTfig1} represents the percentage of correct predictions out of $310$ posterior samples on a single image. $91.8\%$ of these cells have more than $99\%$ confidence. 

Table \ref{MNIST1} shows the performance of our Bayesian approach on the MNIST data. Out of the 310 posterior samples for each image in the test set of 10,000 images considered, the best fit test error represents the mean of the number of erroneous predictions in all samples for all images. In our study, we have obtained a test ensemble accuracy of 99.22 $\%$, which is performance competitive with current state-of-the-art image classification methods.

From table \ref{MNIST1}, we note that previous architectures for MNIST analysis either have a Neural ODE architecture without error estimates \cite{chen2018neural} or do not incorporate a Neural ODE for error estimation \cite{shridhar2019comprehensive}. Incorporated Neural ODEs in our approach, we not only demonstrate a classification performance competitive with current state-of-the-art image classification methods; but also quantify the confidence of our prediction. 
 
\subsection{Bayesian Neural ODE: SGLD}
\begin{figure}
\centering
\begin{tabular}{c}
\subfloat[]{\includegraphics[width=0.5\textwidth]{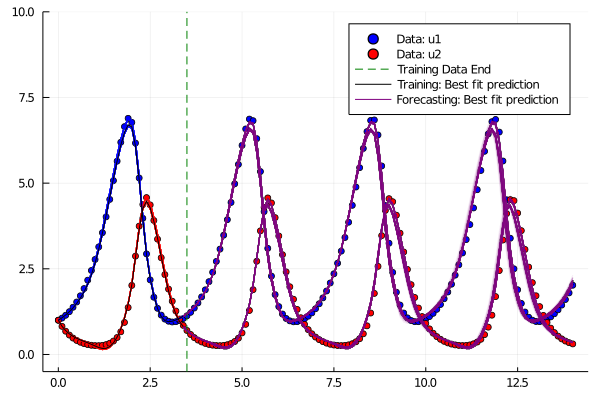}}
\subfloat[]{\includegraphics[width=0.5\textwidth]{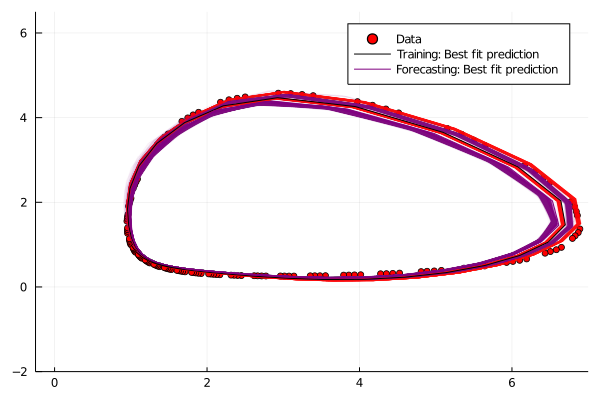}}
\end{tabular}
\caption{Comparison of the Bayesian Neural ODE: SGLD estimation and data for the Lotka Volterra ODE case study shown as (a) time series plots and (b) contour plots.}\label{fig_lvode_sgld}
\end{figure}

Stochastic Gradient Langevin Dynamics (SGLD) is an adaptation, designed to sample from the posterior as the iterations increase, of the usual stochastic gradient descent algorithm. In each iteration, we update our vector $\theta$ of parameters according to the rule
\begin{align}\label{sgld}
    \theta_t &:= \theta_t - \Delta \theta_t \\
    \Delta\theta_t &:= \frac{\epsilon_t}{2}\left(\nabla \log p(\theta_t) + \frac{1}{n}\sum_{i=1}^n \nabla\log p(\mathcal{D}_n\vert \theta)\right) + \eta_t\\
    \eta_t &\sim \mathcal{N}(0, \epsilon_t)
\end{align}
where $\mathcal{D}_n$ are the minibatches the training dataset $\mathcal{D}$ has been split into. $p(\mathcal{D}_n\vert \theta)$ is the likelihood, whose logarithm is equivalent to the loss function, and $p(\theta_t)$ is any priors, also known as regularisation terms, imposed onto the parameters $\theta$. The stepsizes $\epsilon_t$ must follow a decaying scheme which satisfies the conditions \cite{welling2011bayesian}:
\begin{equation}\label{stpsize_conds}
    \sum_{t=1}^\infty \epsilon_t = \infty \quad \sum_{t=1}^\infty \epsilon_t^2 < \infty
\end{equation}
in this article we have chose a polynomial decaying scheme $\epsilon_t = a(b + t)^{-\gamma}$ with $a, b,$ and $\gamma$ as tuneable hyperparameters.

The update scheme for $\theta$ is composed of two stages. In the first stage, where the approximate gradient dominates, we approach the regions with higher mass probability. During the second phase, instead of allowing $\theta$ to converge to a single value, it walks randomly with a predominantly Gaussian noise since the gradient is $\mathcal{O}(\epsilon_t)$ and the Gaussian noise is $\mathcal{O}(\sqrt{\epsilon_t})$. It is on this second stage where it is theoretically guaranteed to converge to the posterior, and hence, we may use this stage to sample parameters from the posterior. 

We now apply SGLD on the Lotka Volterra system in Equations 3-4.
The Lotka Volterra system used in this case has the same parameters as the one used for NUTS. Again, we apply SGLD with $45000$ iterations and sampled
the last $2000$ updates. The hyperparameters used were $a = 0.0025, b = 0.05, \gamma = 0.35$. The neural ODE architecture was again $2$ layers with $50$ neurons and $\textrm{tanh}$ activation. The algorithm took approximately $679$ seconds to run.

From figure \ref{fig_lvode_sgld} we notice a good fit on the training dataset. The Bayesian Neural ODE trained using the SGLD approach has accurately captured the periodicity of the system; and is seen to generalize for a much longer duration than the NUTS sampler.

\subsubsection{Comparison between SGLD and NUTS}
\begin{figure}
\centering
\begin{tabular}{cc}
\subfloat[]{\includegraphics[width=0.45\textwidth]{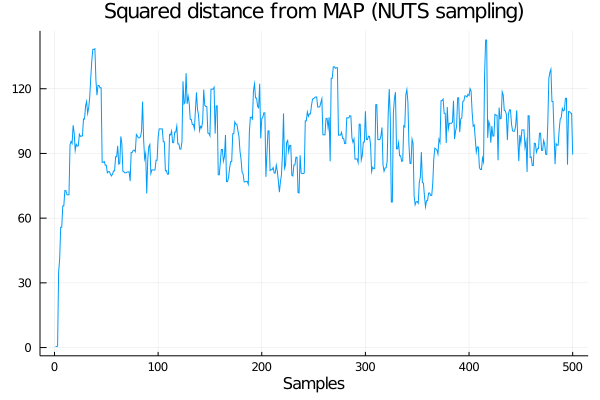}}
\subfloat[]{\includegraphics[width=0.45\textwidth]{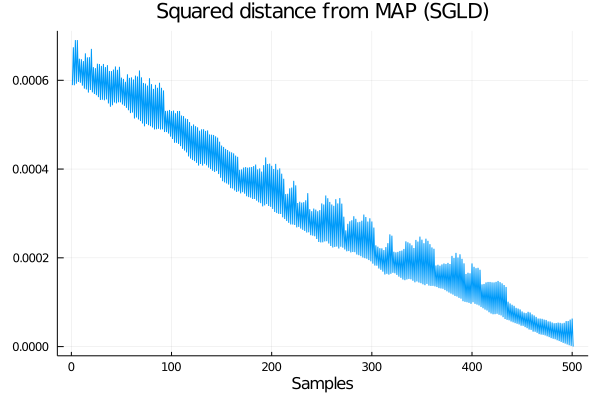}}\\
\end{tabular}
\caption{(a) Figure shows that for the NUTS sample initialized at the MAP point, the sampler quickly jumps away from the MAP point and never returns back. (b) Figure shows that for the SGLD sampler initialized at the MAP, all posteriors samplers are close to the MAP point. }\label{MAP}
\end{figure}

Through figures \ref{spiralfig} and \ref{fig_lvode_sgld}, we note that SGLD generally has a better mean prediction accuracy than NUTS. This can be attributed to the non-convexity/multi-modality of the likelihood function where the MAP point is likely to be in a region with low probability mass, leading the NUTS sampler which uses non-stochastic gradients to not "find" the region surrounding the MAP point in the time of the sampling. The use of stochastic gradients in SGLD seems to have led to a sample much closer to the MAP point and with a much lower mean prediction error. This difference between NUTS and SGLD is illustrated in figure \ref{MAP}, where the distance of the posterior samples from the MAP point is shown. Figure \ref{MAP}a shows that for the NUTS sample initialized at the MAP point, the sampler quickly jumps away from the MAP point and never returns back. Figure \ref{MAP}b shows that for the SGLD sampler, all posteriors samplers are much closer to the MAP point, than the NUTS sampler.

\subsubsection{Application to Neural PDE's}
\begin{figure}
\centering
\begin{tabular}{cc}
\subfloat[]{\includegraphics[width=0.48\textwidth]{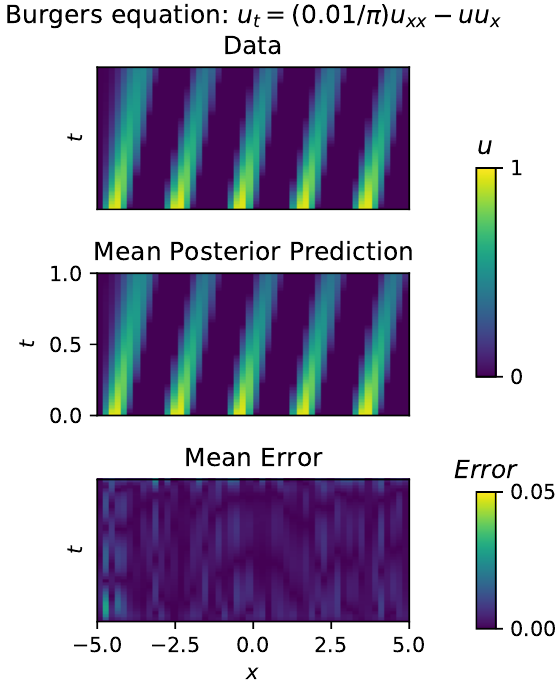}}
\subfloat[]{\includegraphics[width=0.46\textwidth]{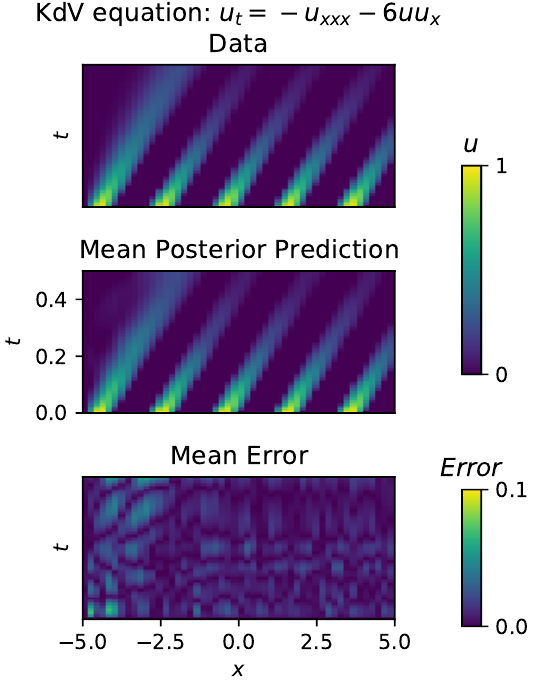}}\\
\end{tabular}
\caption{Figure shows the comparison between the mean posterior prediction for the trained Neural PDE with the true data, and also the resulting error for (a) Example 1: Burgers' equation and (b) Example 2: KdV equation  }\label{PDE_SGLD}
\end{figure}

The SGLD methodology can be readily extended for uncertainty quantification analysis of a Neural PDE. The input vector which varies in space at the initial time ($u(x, 0)$)  can  be propagated through a neural network architecture \textrm{NNODE} with weights $p$, and reformulated as a ODE according to
\begin{equation}
   \frac{\partial u(x, t)}{\partial t} = \textrm{NNODE}(u(x, 0), p)
\end{equation}
The solution of this ODE, $\hat{u}$ will be an $n \times m$ vector where $n$ is the number of time points discretization and $m$ is  the number of space points discretization. Using an L2 loss function based on the difference between $\hat{u}$ and the true data $\bar{u}$, the SGLD methodology outlined in section $2.3$ can be applied in a similar manner to provide the posterior of the neural network parameters $p$, and thus enable uncertainty quantification in the PDE solution.\newline

We illustrate the applicability of SGLD to PDEs using the examples below \newline

\textbf{Example 1: Burgers' equation} \newline
The Burgers' equation is a PDE which shows in wide range of fields including fluid dynamics, gas dynamics and acoustics and is given by
\begin{equation}\label{PDE1}
   \dv{u}{t} = \frac{0.01}{\pi}\dv[2]{u}{x} - u  \dv{u}{x}
\end{equation}

\textbf{Example 2: Korteweg–De Vries (KdV) equation} \newline
The KdV equation is a PDE which shows up in the field of shallow water analysis and is given by
\begin{equation}\label{PDE1}
   \dv{u}{t} = - \dv[3]{u}{x} - 6u \dv{u}{x}
\end{equation}

To generate the true data for these examples, the $x \times t$ space was divided into an $n \times n$ grid with $n = 51$ for both examples. The PDE discretization was performed using the method of lines and then solved using an adapative ODE solver in Julia. \newline

The input vector to the neural PDE architectures in both cases was of size $n \times 1 = 51 \times 1$. We used 2 layers with $10$ neurons in each layer and the relu activation function. Propagation of the input vector to this neural network led to an output of size $n \times n = 51 \times 51$. \newline

We ran the SGLD algorithm with $40000$ iterations and sampled the last $600$ updates. The hyperparameters used were $a = 0.001, b = 0.15, \gamma = 0.05$ for both examples. \newline

Since the SGLD method leads to sampling from the true parameter posterior, we can compare the mean posterior prediction for the trained Neural PDE with the true data and also the resulting error. Figures \ref{PDE_SGLD}a, b shows this comparison and a reasonable agreement is seen. Through this method, we also get estimates of the error in the mean posterior prediction, and not just a deterministic prediction.\newline

The extension of the SGLD method to PDE's further strengthens its validity as a useful Bayesian Neural ODE/PDE method.

\subsection{Bayesian Neural ODE: Variational Inference}
\begin{figure}[h]
\centering
\begin{tabular}{cc}
\subfloat[]{\includegraphics[width=0.45\textwidth]{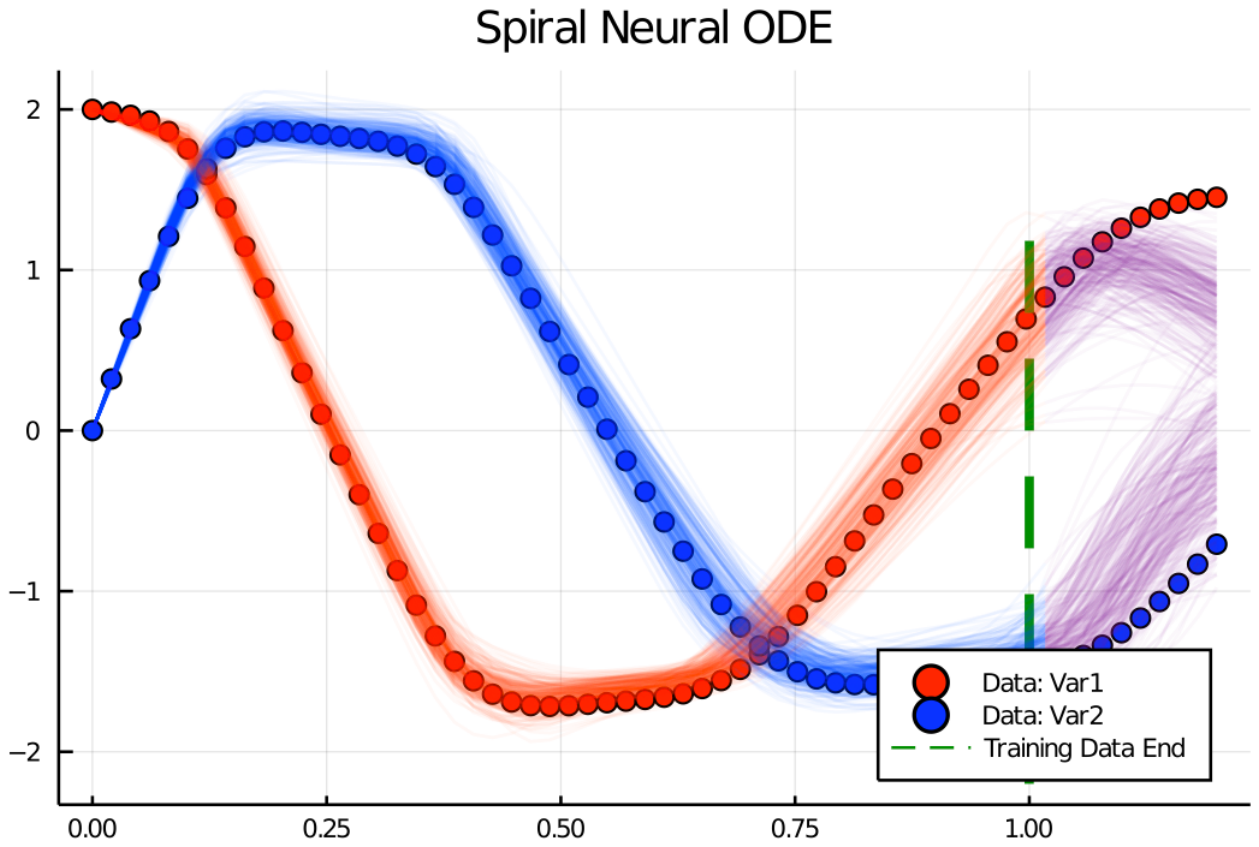}}\\
\subfloat[]{\includegraphics[width=0.38\textwidth]{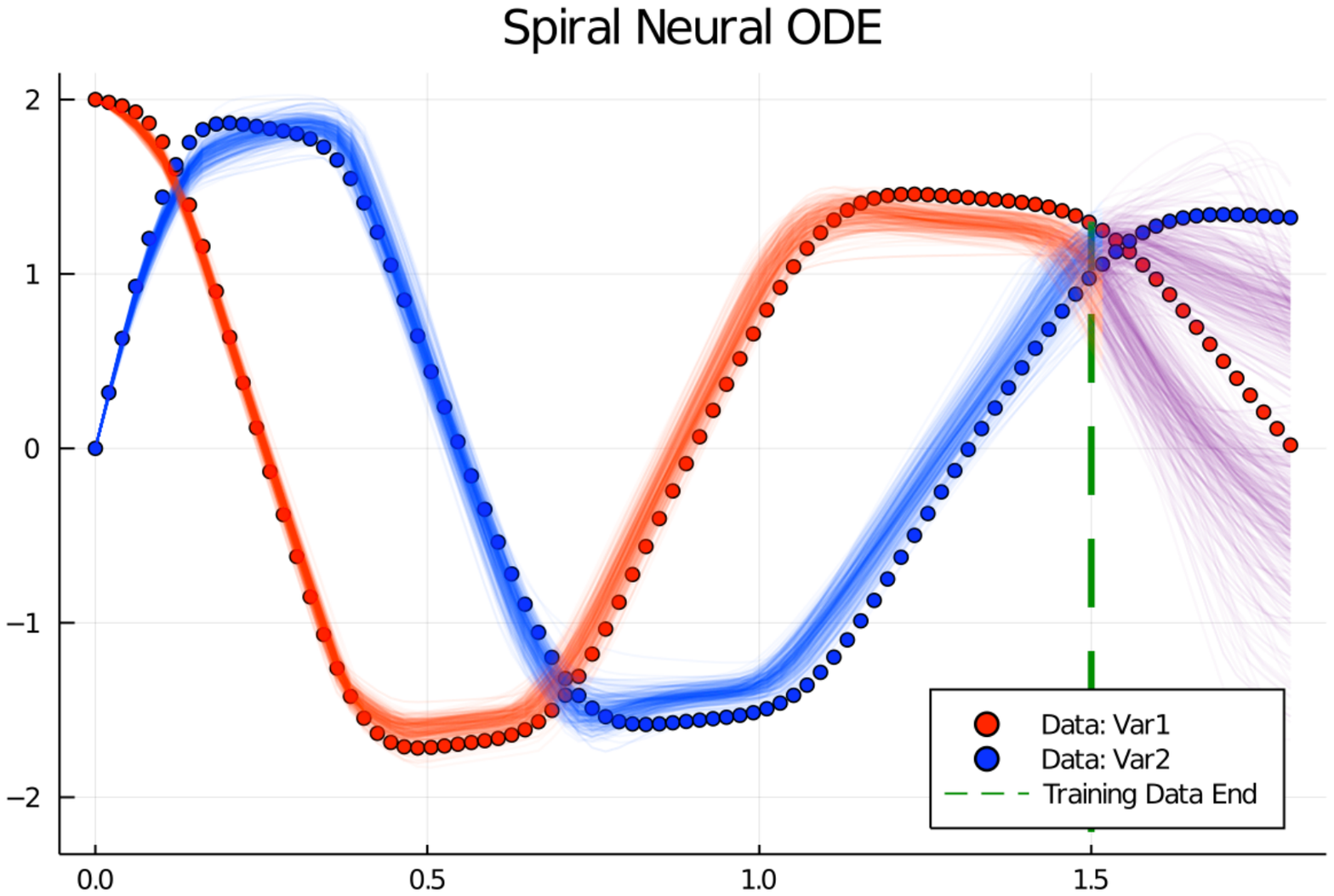}}
\end{tabular}
\caption{For the Spiral ODE example (Equations 1-2), figure shows the retrodiction plots for (a) Variational Inference framework used in the present study and (b) Variational Inference integrated with Normalizing Flow. We can see that integration with normalizing flows used shows marginal improvement over plain Variational Inference with mean field approximation.}\label{VI_spiral_diagnostics}
\end{figure}
Since the last decade, there has been explosive interest in applying variational inference (VI) to text analysis, generative image modeling, and physical/chemical systems analysis \cite{nf1}. VI, an optimization-based approach, generally approximates the posterior much faster than traditional MCMC methods. Variational inference was traditionally used for learning in graphical models (e.g. the sigmoid belief network) \cite{vinew1, vinew2}. However, there have been significant expansions in the methodology and application domains of VI; examples include stochastic variational inference \cite{hoffman2013stochastic1} on large-scale data analysis and black-box variational inference's \cite{ranganath2014black} usage beyond conditionally conjugate models. Through the use of normalizing flows \cite{nf1}, implicit distributions \cite{implicit1, implicit2} and importance-weighted variational autoencoders \cite{weighted1}, the posterior-approximating variational inference family was made more expressive and thus powerful.  

Here, we aim to address whether variational inference methods can be integrated into, and thus further expand, our Bayesian Neural ODEs framework. We use the Turing.jl interface in Julia \cite{turing}. Initially, we define a multivariate normal distribution as the posterior family approximating the true posterior with a mean-field approximation. For maximizing the expected lower bound, we use ADVI (Automatic Differentiation Variational Inference) \cite{kucukelbir2017automatic}, with $10$ samples per step and $5000$ as the upper bound on the number of gradient steps. Figure \ref{VI_spiral_diagnostics}a shows the poor forecasting performance of the Bayesian Neural ODE: Variational Inference framework applied to the Spiral ODE example, compared to HMC methods explored above (figures \ref{spiralfig}, \ref{spiralfig_sghmc}).

\subsubsection{Integration of Normalizing Flows}
One possible reason for the poor forecasting performance of the VI + Neural ODE framework could be due to the posterior family not being expressive enough. To further explore the effects of a powerful posterior family, we look at normalizing flows. 

Through a chain of invertible mappings, normalizing flows transform an initially simple posterior family distribution (such as the multivariate normal in the above example) into an arbitrary complex distribution \cite{nf1}. Considering an initial random variable $z$ with a distribution prescribed by $q(z)$. A bijective mapping function $f$ with inverse $g$, when applied to $z$ results in a new variable $z'$ with distribution given by

\begin{equation}\label{nf1}
q'(z') = q(z) |\textrm{det} \frac{\partial{g}}{\partial{z'}}| = q(z) |\textrm{det} \frac{\partial{f}}{\partial{z}}|^{-1}    
\end{equation}

In their original work, \cite{nf1} demonstrated two types of normalizing flows: planar layer and radial layer. In this study, we have employed the planar layer. The planar layer, parametrized by $u, w, b$ is given by the invertible function

\begin{equation}\label{nf2}
    f(z) = z + u h (w^{T}z + b)
\end{equation}

where $h$ is a differentiable element-wise non-linearity. 
Thus, applying a sequence of maps  $f_{k}$ to an initial density leads to the transformed variable and corresponding density as

 \begin{align}
\label{nf3}
\begin{split}
z_{k} &= f_{k} \circ f_{k-1} \circ \ldots \circ f_{1}(z) 
\\
  \textrm{ln} q_{k}(z_{k}) &= \textrm{ln} q_{0}(z) - \sum_{k = 1}^{K} \textrm{ln} | 1 + u_{k}^{T} \psi_{k} z_{k-1}|
  \end{split}
\end{align}
where $\psi(z) = h' (w^{T}z + b) w$. Application of a planar flow to a standard Gaussian distribution leads to flexible expansions/contractions along hyperplanes and thus makes the base distribution a lot more expressive.

In the present study, the base distribution was initialized as a multivariate normal distribution with mean as a randomly initialized vector with length governed by the number of parameters of the neural ODE. We transformed this base distribution using a composition of 2 such planar layers described in Equation (\ref{nf2}); with $z$ being the parameters of the neural ODE. 

Figure \ref{VI_spiral_diagnostics}b shows that through the inclusion of normalizing flows, the estimation performance of the Bayesian Neural ODE: Variational Inference object is marginally better compared to figure \ref{VI_spiral_diagnostics}a. For the Variational Inference experiment we used  During experiments with different configurations such as Neural Network size, time span of solution and initial weights for the Neural Network used as well for the Normalizing Flow layers had considerable effect on the training performance and forecasting capability of the model. There are quite a lot of open questions that need to be addressed to ensure suitability of Variational Inference with Neural ODEs and will be part of future work.

\subsection{Bayesian Neural UDE: SGLD and PSGLD}
In this section, we aim to demonstrate a viable method for the probabilistic quantification of epistemic uncertainties via a hybrid machine-learning and mechanistic-model-based technique.

\begin{figure*}
\centering
\begin{tabular}{c}
\subfloat[]{\includegraphics[width=0.35\textwidth]{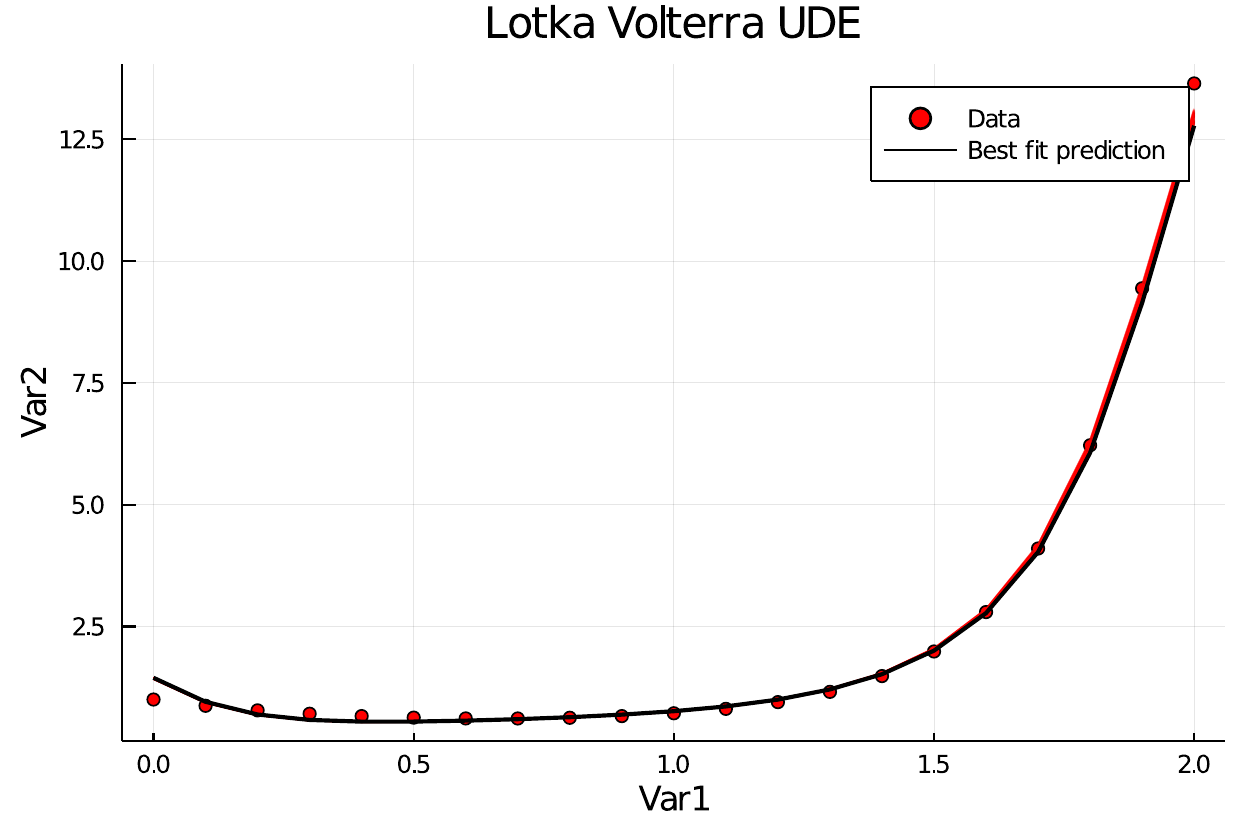}}
\subfloat[]{\includegraphics[width=0.35\textwidth]{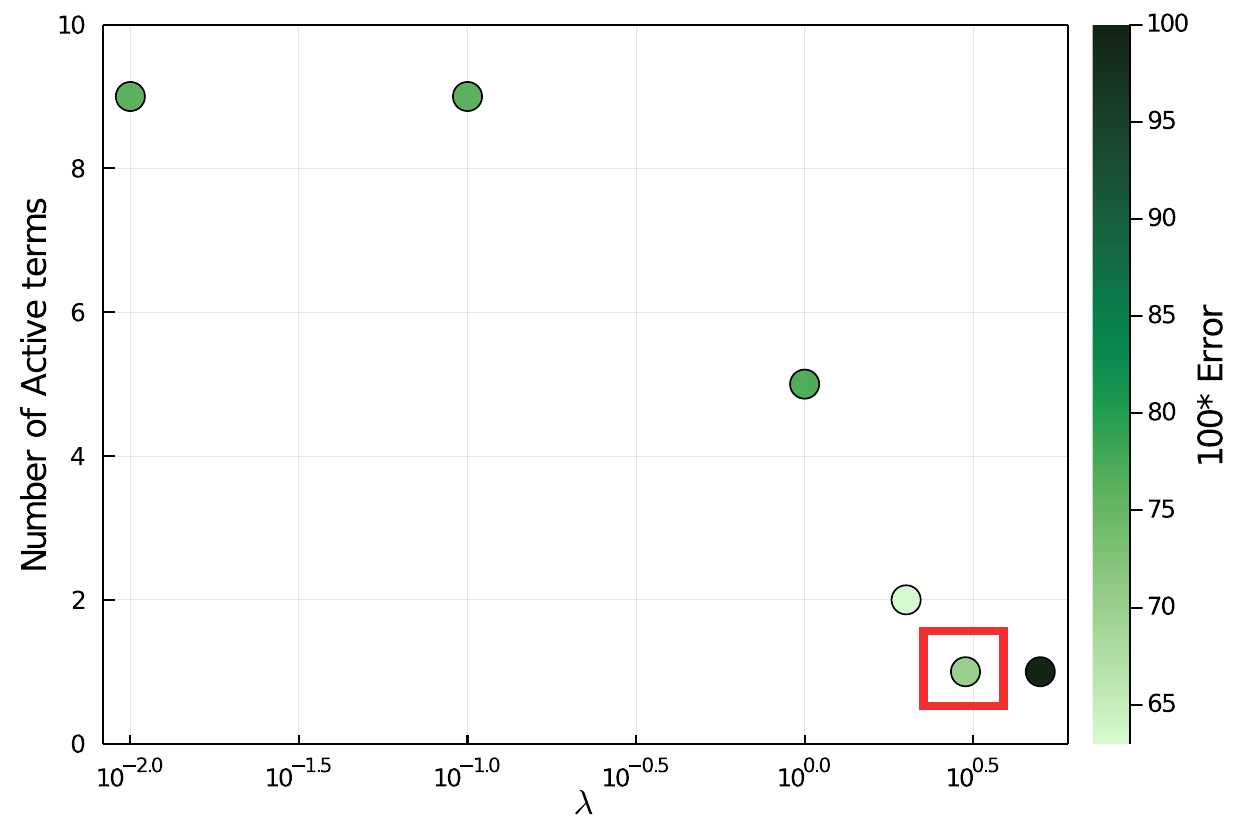}}
\end{tabular}
\caption{Bayesian Neural UDE estimation is demonstrated for the Lotka Volterra example with a missing term as shown in (\ref{lv_ude}); using the PSGLD approach. (a) Comparison of the recovered missing term and the actual term and (b) Sparsity plot using the STRRidge algorithm. The highlighted box shows the optimal point which gives the sparsest solution (1 term) with a low error. This plot is seen to be the same for $100$ trajectories considered in the sampling phase.}\label{lvudefig}
\end{figure*}

More efficient training of deep neural networks is achieved by using a preconditioned matrix $G(\theta)$ in the gradient update step of Equation 15 of the SGLD method \cite{li2015preconditioned}. Combining this with an adaptive step size method like RMSprop leads to much faster sampling than the standard SGLD approach, as outlined in the Preconditioned SGLD with RMSprop algorithm by \cite{li2015preconditioned}. We demonstrate the use of this algorithm in this section, where standard SGLD failed to converge to the true posterior.\newline

\subsubsection{Application to a predator-prey model}
As outlined by \cite{rackauckas2020universal}, universal differential equations (UDE's) can be used to recover missing terms of governing equations describing dynamical systems. As an example, we look at the Lotka Volterra system with a missing term denoted by $M(u_{1}, u_{2})$ in the first variable derivative as

\begin{align}\label{lv_ude}
    \frac{du_{1}}{dt} & = - \alpha u_{1} - M (u_{1}, u_{2})\\
    \frac{du_{2}}{dt} & = - \delta u_{2} + \gamma u_{1} u_{2}\\
\end{align}

\begin{table*}[h]
  \caption{Bayesian Neural UDE: Recovery of dominant terms for the Lotka Volterra example, as the sparsity parameter $\lambda$ is varied. Highlighted row shows the sparsity parameter with the lowest positive AIC score.}
  \label{table1}
  \centering
  \begin{tabular}{llllll}
    \toprule
    \cmidrule(r){1-2}
    $\lambda$  & Number of    & Dominant terms    & Error & Mean & $\%$ \\ & Active terms & & & AIC score & sampled \\
    \midrule
    0.01 & 9  & $u_{1}^{2}, u_{2}^{2}, u_{1} u_{2}$   & 0.765 & 40.4 & 100\\  & & $u_{1}^{2} u_{2}^{2}, u_{1}^{2} u_{2}, u_{2}^{2} u_{1}$ & & \\ & & $u_{1} u_{2}$, const & & \\
    0.1 & 9  & $u_{1}^{2}, u_{2}^{2}, u_{1} u_{2}$   & 0.764 & 35 & 100\\  & & $u_{1}^{2} u_{2}^{2}, u_{1}^{2} u_{2}, u_{2}^{2} u_{1}$ & & \\ & & $u_{1} u_{2}$, const & & \\
    1 & 5  & $u_{1}^{2}, u_{2}^{2}, u_{2}$   & 0.764 & 21.6 & 100\\  & & $u_{1}^{2} u_{2}, u_{1} u_{2}$ & & \\ 
    2 & 2  & $u_{1}^{2} u_{2}, u_{1} u_{2}$   & 0.634 & 7.2 & 100\\ 
        \midrule
   \textbf{3} & \textbf{1}  & \textbf{$u_{1} u_{2}$}   & \textbf{0.7} & \textbf{4.1} & \textbf{100}\\ 
       \midrule
    5 & 1  & $u_{1}^{2} u_{2}$   & 2.49 & -1 & 100\\ 

    \bottomrule
  \end{tabular}
\end{table*}

Using the PSGLD method outlined in \cite{rackauckas2020universal}, we trained $M_{\theta} (u_{1}, u_{2})$ as a neural network to optimize the weights $\theta$; and recover the missing term time series. We sampled from the last 100 updates of the converged sampler. Figure \ref{lvudefig}a shows that the recovered time series $M_{\theta}(u_{1}, u_{2})$ from 100 trajectories matches very well with the actual term $M_{\theta}(u_{1}, u_{2}) = u_{1} u_{2}$. These optimized parameter space for all $100$ trajectories lies very close to each other, indicating that the PSGLD method indeed converges and then subsequently samples closer to the true posterior. 

Subsequently, we used a sparse regression technique called Sequential Thresholded Ridge Regression (STRRidge) algorithm \cite{brunton2016discovering} on the neural network output to reconstruct the missing dynamical equations for $100$ trajectories of the sampled parameter space. The STRRidge algorithm has a tunable sparsity parameter $\lambda$ to control the sparsity of the obtained dominant terms. Optimally, we would want the sparsest solution with the least possible error. 

Figure \ref{lvudefig}b shows the variation of the number of terms recovered by the STRRidge algorithm with the sparsity parameter, $\lambda$; which shows a decreasing trend as expected. The colorbar indicates the error between the sparse recovered solution and the neural network output $M_{\theta}(u_{1}, u_{2})$. It can be seen that the highlighted box indicates the optimal point which has the sparsest solution (1 term) with a very low error. This point also corresponds to the lowest positive AIC score (Akaike Information Criteria) \cite{aic2, aic1} which strives to minimize the model error as well as its complexity (shown in Table \ref{table1}). Along with showing the AIC score as function of the sparsity parameter $\lambda$, table \ref{table1} also shows the dominant terms as the sparsity parameter $\lambda$ is varied.

This optimal solution has the quadratic form $\sim u_{1} u_{2}$, for all $100$ trajectories indicating that Bayesian Neural UDE approach recovers the correct solution for all sampled trajectories. Out of the $100$ models sampled for the optimal sparsity parameter, the model with the lowest AIC score was found to be $M(u_{1}, u_{2}) = 0.96 u_{1} u_{2}$, which is very close to the true solution $M(u_{1}, u_{2}) = u_{1} u_{2}$. 
\subsubsection{Application to an epidemiology model}
\begin{figure}
\centering
\begin{tabular}{c}
\subfloat[]{\includegraphics[width=0.35\textwidth]{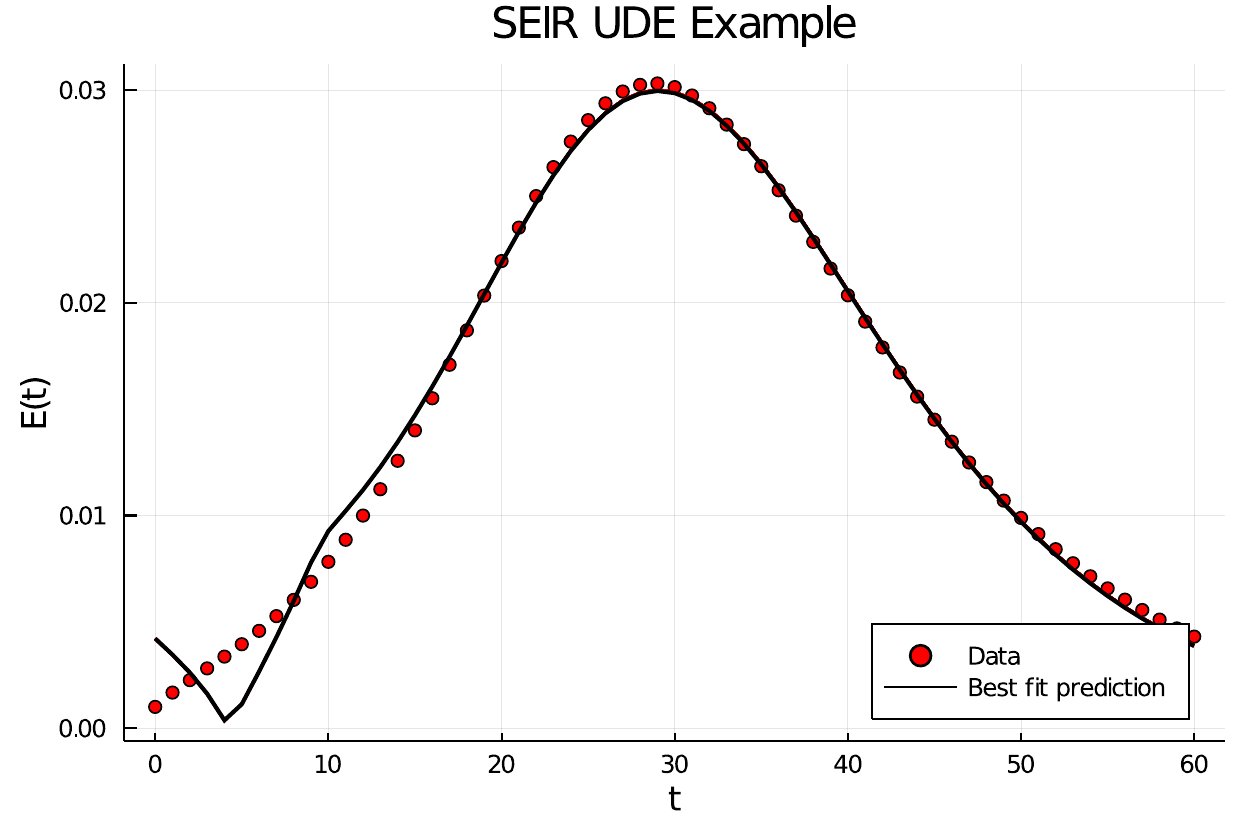}}
\end{tabular}
\caption{Bayesian Neural UDE estimation is demonstrated for the SEIR example with a missing term as shown in (\ref{seir_ude2}); using the PSGLD approach. (a) Comparison of the recovered missing term and the actual term shown for $100$ trajectories considered in the sampling phase.}\label{seirudefig}
\end{figure}

As another example to test our approach, we consider a SEIR epidemiological model. The SEIR is a compartment based model which models transfer of population between four compartments: Susceptible, Exposed, Infected and Recovered using the following set of equations:

\begin{align}\label{seir_ude}
    \frac{dS}{dt} & = - \beta S I \\
    \frac{dE}{dt} & = \beta S I - \sigma E \\
    \frac{dI}{dt} & = \sigma E - \gamma I \\
    \frac{dR}{dt} & = \gamma I \\
\end{align}

In Equation \ref{seir_ude}, we will try to infer the exposure term using Bayesian Neural UDE: PSGLD approach. Thus, we will try to learn $M(E, I) = \sigma E$ in the following system of equations

\begin{align}\label{seir_ude2}
    \frac{dS}{dt} & = - \beta S I \\
    \frac{dE}{dt} & = \beta S I - M(E, I) \\
    \frac{dI}{dt} & = \sigma E - \gamma I \\
    \frac{dR}{dt} & = \gamma I \\
\end{align}
Using the PSGLD method outlined in \cite{rackauckas2020universal}, we trained $M_{\theta} (E, I)$ as a neural network to optimize the weights $\theta$; and recover the missing term time series. We sampled from the last 100 updates of the converged sampler. Figure \ref{seirudefig}a shows that the recovered time series $M_{\theta}(E, I)$ from 100 trajectories matches very well with the actual term $M_{\theta}(E, I) = \sigma E$.

Subsequently, we applied the STRRidge algorithm to recover the symbolic equations for the missing terms, with the sparsity parameter $\lambda$ ranging from $0.005 - 0.5$. The model for which the lowest AIC score was obtained, was found to contain just one dominant term ($E$) and with the symbolic form $M(E, I) = 0.099 E$ compared to the ground truth data of $M(E, I) = 0.1E$, for all $100$ trajectories sampled. 

\subsection{Application to PDE's: Wave propagation}

\begin{figure*}
\centering
\begin{tabular}{ccc}
\subfloat[]{\includegraphics[width=0.35\textwidth]{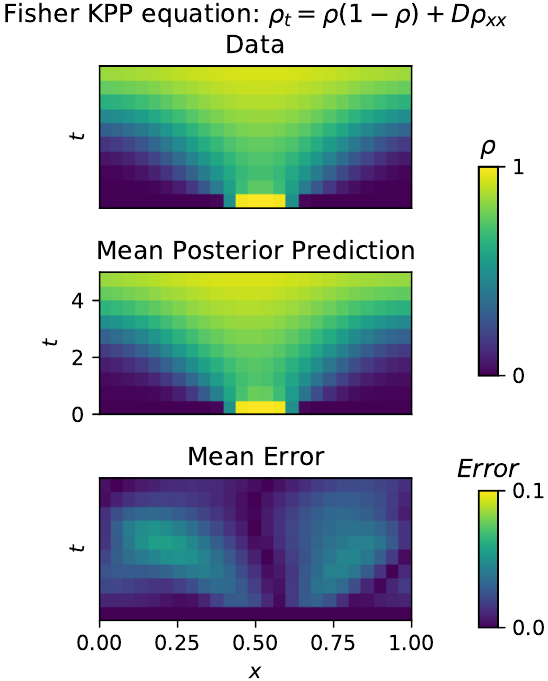}}
\subfloat[]{\includegraphics[width=0.6\textwidth]{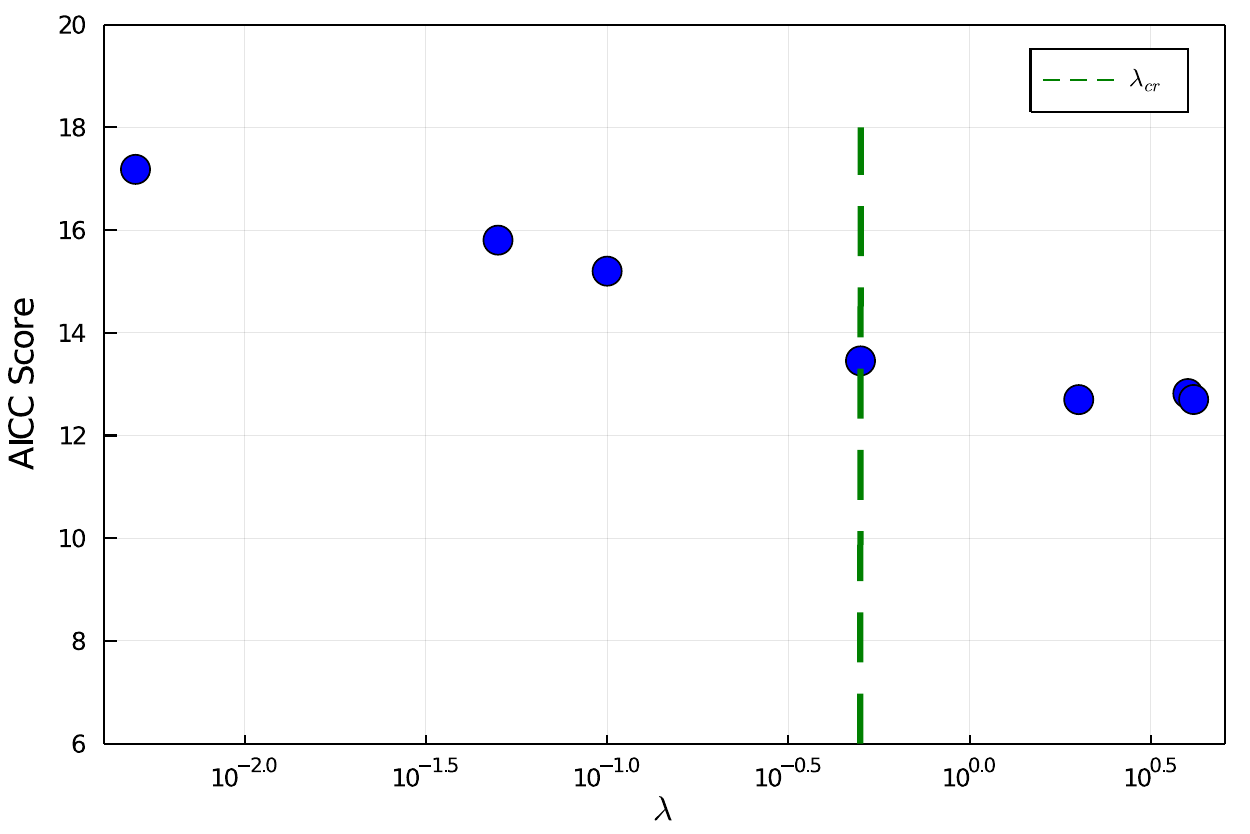}}\\
\subfloat[]{\includegraphics[width=0.35\textwidth]{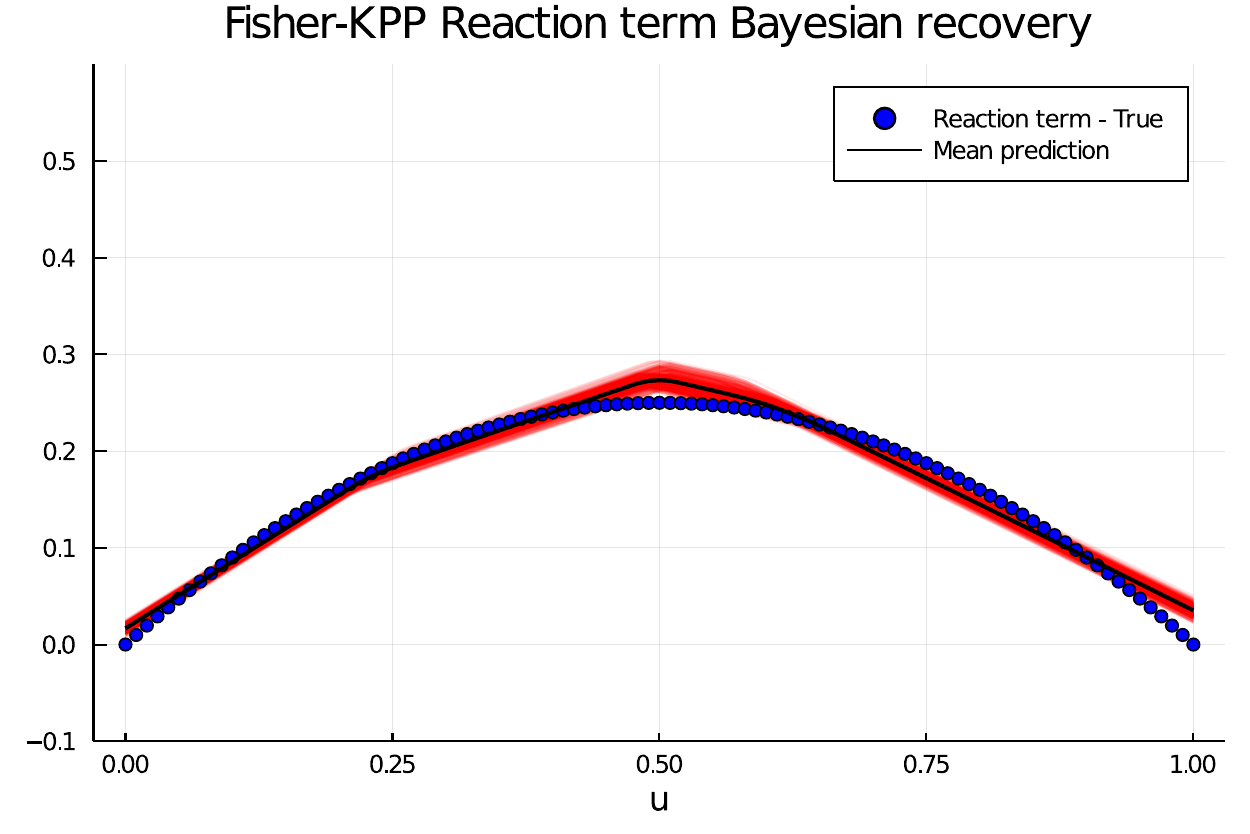}}
\subfloat[]{\includegraphics[width=0.35\textwidth]{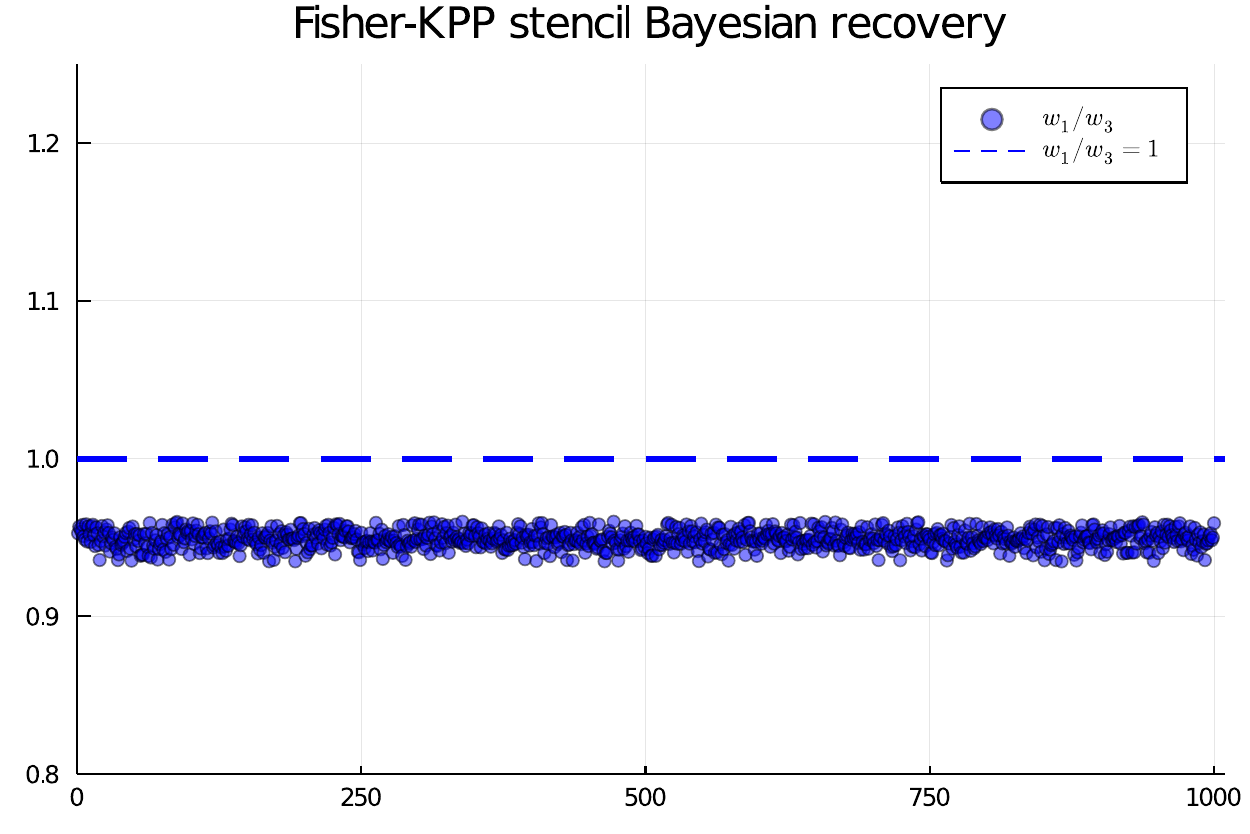}}
\subfloat[]{\includegraphics[width=0.35\textwidth]{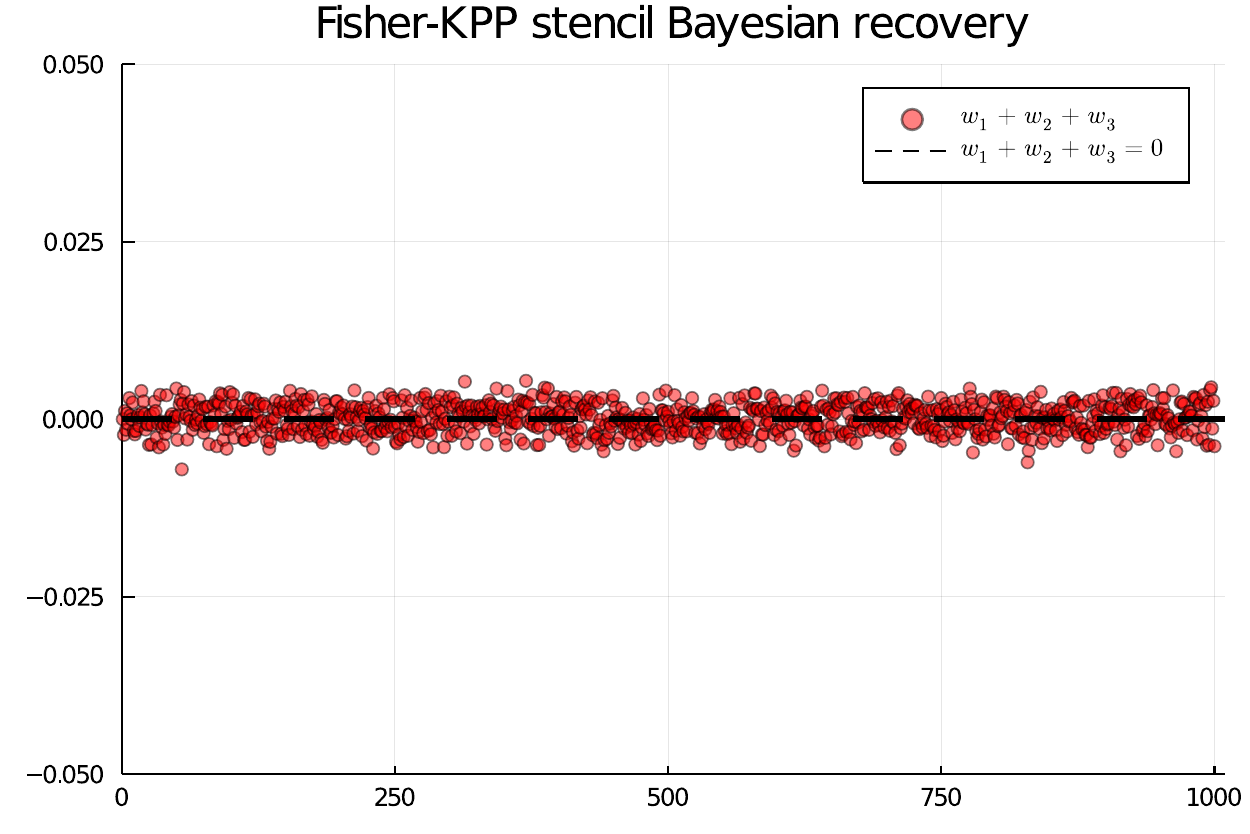}}\\
\end{tabular}
\caption{Bayesian Neural UDE estimation is demonstrated for the Fisher-KPP PDE example with a missing term as shown in (\ref{Eqn:UPDE-Fisher}); using the PSGLD approach. Figures show comparison between: (a) Training data and the mean of $500$ recovered posterior solutions, (c) True reaction term and the posterior recovered term, (d, e) Posterior recovered weights for the convolutional filter and the canonical stencil [$1, -2, 1$] for the one-dimensional Laplacian. (b) The variation of the sparsity parameter $\lambda$ in the STRRidge algorithm, with the obtained AIC score}\label{fkpp}
\end{figure*}

\begin{table*}[h]
  \caption{Bayesian Neural UDE: Recovery of the quadratic reaction term for the Fisher-KPP equation. Results are shown for the sparsity parameter $\lambda = \lambda_{cr} = 0.5$ for which the AIC score begins to show a plateau (figure \ref{fkpp}b). Results are shown for $1000$ posterior samples }
  \label{tablekpp}
  \centering
  \begin{tabular}{llll}
    \toprule
    \cmidrule(r){1-2}
    $\lambda_{cr}$  & Number of    & Dominant terms   & $\%$ of \\ & Active terms &   & samples \\
    \midrule
    0.5 & 2  & $\rho, \rho^{2}$   & 73 \\  
    0.5 & 3  & $\rho, \rho^{2}, \rho^{3}$   & 27\\ 

    \bottomrule
  \end{tabular}
\end{table*}

To illustrate the probabilistic system identification using Bayesian Neural UDEs, we consider a spatio-temporal system governed by the one-dimensional Fisher-KPP PDE
\begin{equation}\label{Eqn:Fisher-KPP}
\frac{\partial \rho}{\partial t} = r\rho(1-\rho) + D \frac{\partial^2 \rho}{\partial x^2}, 
\end{equation}
with $x\in[0,1]$, $t\in[0,T]$, and periodic boundary condition $\rho(0,t)=\rho(1,t)$. Here $\rho$ represents population density of a species, $r$ is the local growth rate and $D$ is the diffusion coefficient. Such reaction-diffusion equations appear in diverse physical, chemical and biological problems \cite{Grindrod1996}.
To learn the generated data, we define the UPDE:
\begin{equation}\label{Eqn:UPDE-Fisher}
\rho_t = \text{NN}_\theta(\rho) + \hat{D}\,\text{CNN}(\rho),
\end{equation}
where $\text{NN}_\theta$ is a neural network representing the local growth term. The derivative operator is approximated as a convolutional neural network $\text{CNN}$, a learnable arbitrary representation of a stencil while treating the coefficient $\hat{D}$ as an unknown parameter fit simultaneously with the neural network weights. We encode in the loss function extra constraints to ensure the learned equation is physically realizable, i.e. the derivative stencil must be conservative (the coefficients sum to zero). \newline

We trained the neural network $\text{NN}_\theta$ and the convolutional neural network $\text{NN}_\theta$ using the PSGLD method outlined above. We sample from the last $1000$ updates of the converged posterior. Figure \ref{fkpp}a shows that the mean of the posterior recovered solutions shows a good match with the training data. Figure \ref{fkpp}c shows the bayesian recovery of the reaction term, parameterized by $\text{NN}_\theta$, which shows a quadratic form for all $1000$ posterior samples with mean value lying close to the true quadratic reaction term. Figure \ref{fkpp}d, e shows the posterior recovered convolutional filter $[w_{1}, w_{2}, w_{3}]$. From these two plots, we see that the posterior for these weights lies very close to the canonical stencil [$1, -2, 1$] for the one-dimensional Laplacian.\newline

Subsequently, we applied the STRRidge algorithm to recover the symbolic equations for the reaction term $r \rho (1 - \rho)$, with the sparsity parameter $\lambda$ ranging from $O(1e-3) - O(1)$. The variation of $\lambda$ with the obtained AIC score is shown in figure \ref{fkpp}b. We see the critical $\lambda$ for which the AIC score starts showing a plateau is marked as $\lambda_{cr} = 0.5$ and shown as a dotted line in figure \ref{fkpp}b. We used the STRRidge algorithm using this value of the sparsity parameter $\lambda$ for $1000$ recovered posterior samples. In table \ref{tablekpp}, we see that $73 \%$ of these posterior samples show the dominant symbolic terms to be $\rho, \rho_{2}$, which matches the true quadratic reaction term form. For the remaining $27\%$, the terms recovered are seen to be $\rho, \rho^{2}, \rho^{3}$.

\subsection{Application to PDE's: Climate models}

\begin{figure*}
\centering
\begin{tabular}{cc}
\subfloat[]{\includegraphics[width=0.45\textwidth]{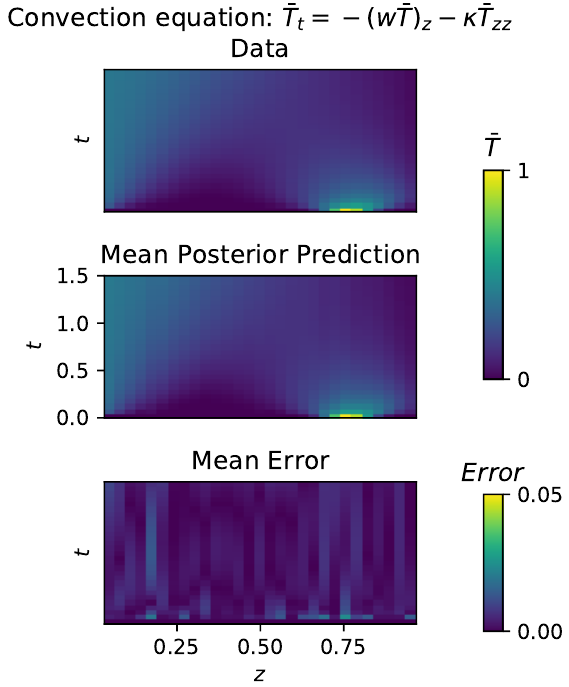}}\\
\subfloat[]{\includegraphics[width=0.45\textwidth]{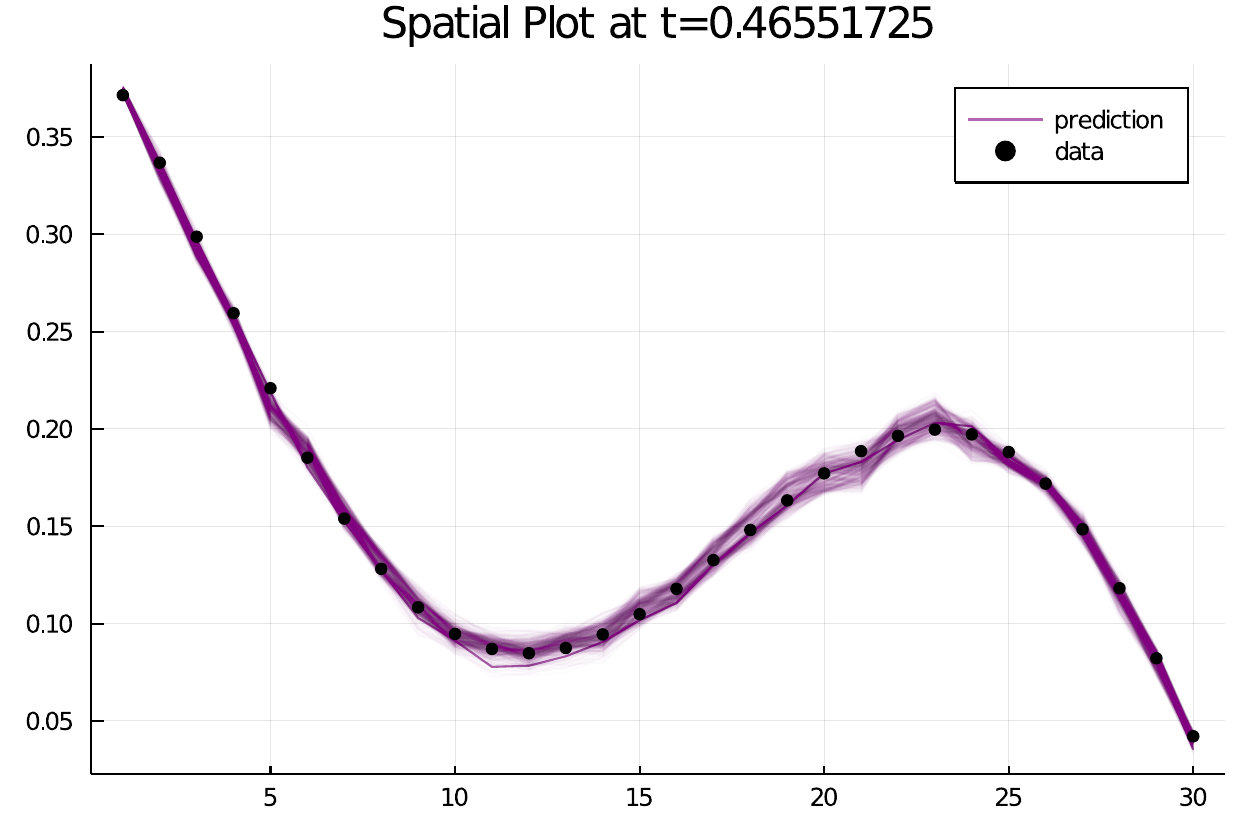}}
\subfloat[]{\includegraphics[width=0.45\textwidth]{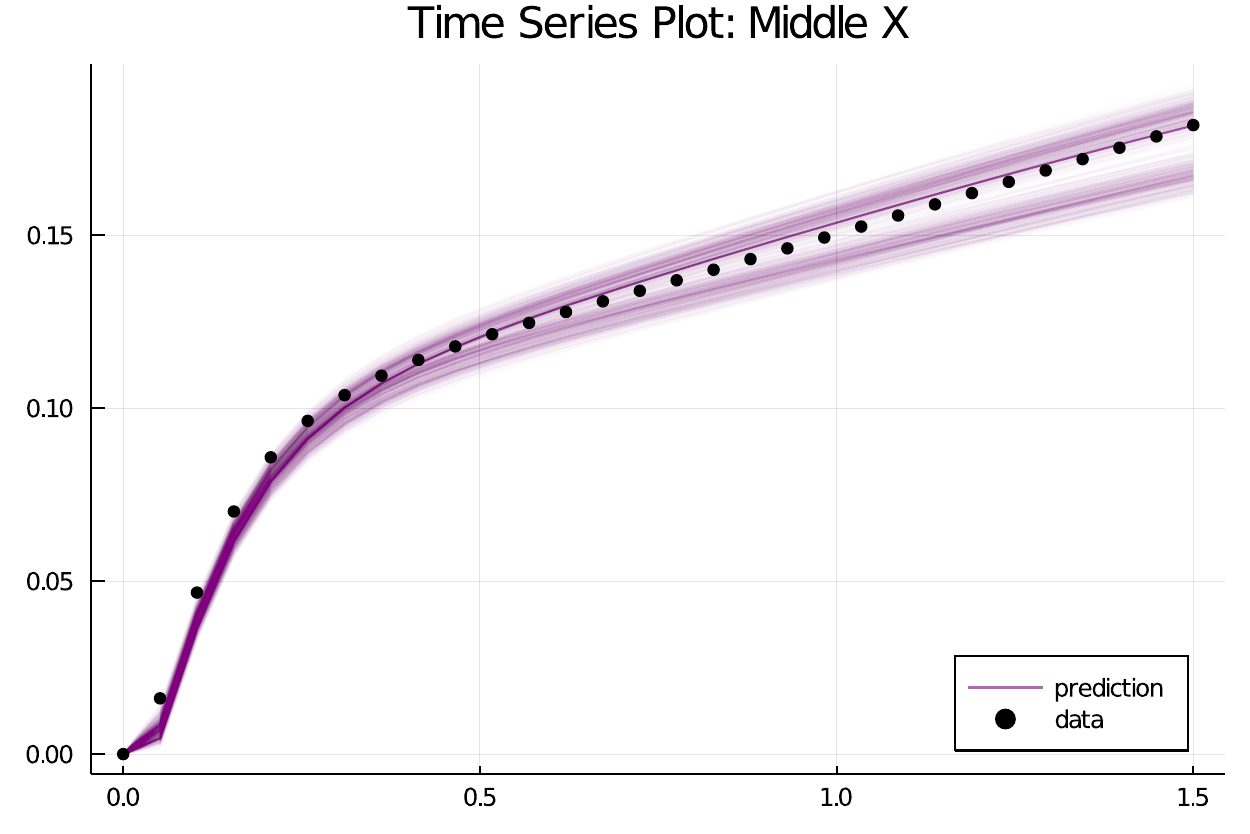}}\\
\end{tabular}
\caption{Bayesian Neural UDE estimation using the PSGLD approach. is demonstrated for the on Eddy Model parametrizations used in Climate models. Figures show comparison between: (a) Training data and the mean of $500$ recovered posterior solutions, (b) Temporal slice of the training data and the recovered posterior and (c) Spatial slice of the training data and the recovered posterior.}\label{eddy}
\end{figure*}

As an example of directly accelerating existing scientific workflows, we focus on the Boussinesq equations \cite{CUSHMANROISIN201199}. The Boussinesq equations are a system of 3+1-dimensional partial differential equations acquired through simplifying assumptions on the incompressible Navier-Stokes equations, represented by the system:

\begin{equation}
\begin{aligned}
\nabla \cdot {\bf u} &= 0, \\
\frac{\partial {\bf u}}{\partial t} + ({\bf u} \cdot \nabla){\bf u} &= -\nabla p + \nu \nabla^2 {\bf u} + b\hat{z}, \\
\frac{\partial T}{\partial t} + {\bf u} \cdot \nabla T &= \kappa \nabla^2 T,
\end{aligned}
\end{equation}
where ${\bf u} = (u,v,w)$ is the fluid velocity, $p$ is the kinematic pressure, $\nu$ is the kinematic viscosity, $\kappa$ is the thermal diffusivity, $T$ is the temperature, and $b$ is the fluid buoyancy. We assume that density and temperature are related by a linear equation of state so that the buoyancy $b$ is only a function $b = \alpha g T$ where $\alpha$ is the thermal expansion coefficient and $g$ is the acceleration due to gravity.

This system is commonly used in climate modeling, especially for modeling the ocean \cite{griffies2008formulating,CUSHMANROISIN201199} in a multi-scale model that approximates these equations by averaging out the horizontal dynamics $\overline{T}(z,t) = \iint T(x,y,z,t) \, dx \, dy$ in individual boxes. The resulting approximation is a local advection-diffusion equation describing the evolution of the horizontally-averaged temperature $\overline{T}$:

\begin{equation}
\frac{\partial \overline{T}}{\partial t} + \frac{\partial \overline{wT}}{\partial z} = \kappa \frac{\partial^2 \overline{T}}{\partial z^2}.
\end{equation}
This one-dimensional approximating system is not closed since $\overline{wT}$ is unknown. Common practice closes the system by manually determining an approximating $\overline{wT}$ from ad-hoc models, physical reasoning, and scaling laws. However, we can utilize a UDE-automated approach to learn such an approximation from data. Let 
\begin{equation}
\overline{wT} = U_\theta\left(P,\overline{T},\frac{\partial \overline{T}}{\partial z}\right)
\end{equation}
where $P$ are the physical parameters of the Boussinesq equation at different regimes of the ocean, such as the amount of surface heating or the strength of the surface winds \cite{CVMix}. We can accurately capture the non-locality of the convection in this term by making the UDE a high-dimensional neural network. \newline

Data was generated from the diffusion-advection equations using the missing function $\overline{wT} = \cos(\sin(T^3)) + \sin(\cos(T^2))$.
Similar to the above examples, we can train this neural network for the $\overline{wT}$ term using the PSGLD approach and obtain bayesian estimates of the recovered solution using the weight posteriors of this neural network. \newline

Figure \ref{eddy}a shows that the mean of the posterior recovered solutions shows a good match with the training data. Figure \ref{eddy}b, c shows the temporal and spatial slices of the training data respectively, compared with the bayesian recovered solutions for $500$ samples. We can see that we can not only recover the correct spatial and temporal variations in the temperature data but also capture the uncertainty associated with the predictions.\newline

Thus, the model discovery for both the predator-prey example and the epidemiological model is robust to uncertainty. The Bayesian Neural UDE framework is thus an added arsenal to the recently demonstrated methods on bayesian system identification \cite{atkinson2020bayesian, yang2020bayesian}. Future work will further identify the relationship between model uncertainties and probabilistic automated discovery.
\section{Conclusion and Future Work}
We have shown that Bayesian learning frameworks can be integrated with Neural ODE’s to quantify the uncertainty in the weights of a Neural ODE, using three sampling methods: NUTS, SGHMC and SGLD. Bayesian Neural ODEs with SGLD sampling is seen to provide better prediction accuracy and less bias from the MAP than NUTS sampling, possibly due to non-convexity/multi-modality of the likelihood function where the MAP point is likely to be in a region with low probability mass. However, a better understanding of why the two algorithms differ and how they compare to other algorithms is needed.\newline

In addition, using a novel architecture which integrates convolution layers and Neural ODEs with the SGHMC framework; we demonstrate a test ensemble accuracy of $99.22\%$ which is comparable with the accuracy of state-of-the-art image classification methods. \newline

Subsequently, for the first time, we demonstrate the integration of Neural ODEs with Variational Inference. We show that when variational inference is combined with normalizing flows, it leads to a good prediction and estimation performance on physical systems; thus leading to a potentially powerful Bayesian Neural ODE object.\newline

Finally, considering the problem of recovering missing terms from a dynamical system using universal differential equations (UDEs); we demonstrate the Bayesian recovery of missing terms from dynamical systems for (a) a predator-prey model and (b) an epidemiological model. \newline

Currently, it is observed that Bayesian learning of Neural ODEs is computationally expensive for large datasets. Therefore, more work is needed to evaluate, understand and improve the convergence of various approximate Bayesian inference and MCMC algorithms in the context of Neural ODEs. Another research direction we plan to delve into further is Bayesian Neural SDE's and their applicability to physical systems and large scale machine learning datasets.

\section{Code Availability}
All codes for SGHMC, Bayesian Neural UDE are publicly available at
\texttt{https://github.com/RajDandekar/MSML21\_BayesianNODE}\newline
and codes for the Variational Inference Neural ODE object is available at
\texttt{https://github.com/mohamed82008/ \newline BayesNeuralODE.jl}

\section{Acknowledgements}
This material is based upon work supported by the National Science Foundation under grant no. OAC-1835443, grant no. SII-2029670, grant no. ECCS-2029670, grant no. OAC-2103804, and grant no. PHY-2021825. We also gratefully acknowledge the U.S. Agency for International Development through Penn State for grant no. S002283-USAID. The information, data, or work presented herein was funded in part by the Advanced Research Projects Agency-Energy (ARPA-E), U.S. Department of Energy, under Award Number DE-AR0001211 and DE-AR0001222. We also gratefully acknowledge the U.S. Agency for International Development through Penn State for grant no. S002283-USAID. The views and opinions of authors expressed herein do not necessarily state or reflect those of the United States Government or any agency thereof. This material was supported by The Research Council of Norway and Equinor ASA through Research Council project "308817 - Digital wells for optimal production and drainage". Research was sponsored by the United States Air Force Research Laboratory and the United States Air Force Artificial Intelligence Accelerator and was accomplished under Cooperative Agreement Number FA8750-19-2-1000. The views and conclusions contained in this document are those of the authors and should not be interpreted as representing the official policies, either expressed or implied, of the United States Air Force or the U.S. Government. The U.S. Government is authorized to reproduce and distribute reprints for Government purposes notwithstanding any copyright notation herein.

\bibliography{neurips}

\bibliographystyle{plain}

\end{document}